\documentclass{article}

\usepackage[accepted]{icml2024}

\usepackage{microtype}
\usepackage{graphicx}
\usepackage{subfigure}
\usepackage{booktabs} 
\usepackage{tikz}
\usepackage{multirow, makecell}
\usetikzlibrary{fadings}
\usetikzlibrary{shapes,snakes}
\tikzset{>=latex}
\usepackage{pgfplots}
\usepackage{float}

\tikzset{
  dots/.style args={#1per #2}{%
    line cap = round,
    dash pattern = on 0 off #2/#1
  }
}

\definecolor{red1}{RGB}{190,20,25}
\definecolor{red2}{RGB}{210,60,40}
\definecolor{blue1}{RGB}{25,100,170}
\definecolor{blue2}{RGB}{75,150,200}
\definecolor{orange1}{RGB}{200,120,30}
\definecolor{orange2}{RGB}{255,170,100}
\definecolor{green1}{RGB}{15,110,60}
\definecolor{green2}{RGB}{120,200,125}
\definecolor{violet1}{RGB}{100,70,150}
\definecolor{violet2}{RGB}{150,150,200}
\definecolor{grey1}{RGB}{100, 100, 100}
\definecolor{grey2}{RGB}{120, 120, 120}
\definecolor{grey3}{RGB}{50, 50, 50}
\definecolor{betterpink}{RGB}{255, 20, 147}

\usepackage[colorlinks=true, allcolors=blue]{hyperref}

\newcommand{\mes}[2]{#1$_{\color{gray} \scriptscriptstyle \pm #2}$}

\newcommand{\defeq}{\vcentcolon=}
\newcommand{\laweq}{\overset{d}{=}}

\newcommand*\Eb[2][]{\mathbb{E}_{#1}\left[ #2 \right]}

\usepackage{amsmath}
\usepackage{amssymb}
\usepackage{mathtools}
\usepackage{amsthm}
\usepackage{bm}

\usepackage[capitalize,noabbrev]{cleveref}

\theoremstyle{plain}
\newtheorem{theorem}{Theorem}[section]
\newtheorem{proposition}[theorem]{Proposition}

\theoremstyle{definition}
\newtheorem{definition}[theorem]{Definition}

\theoremstyle{remark}

\usepackage[textsize=tiny]{todonotes}

\graphicspath{{figs/}}

\newcommand*{\drm}{\mathrm{d}}
\newcommand*{\ve}[1]{{\boldsymbol{#1}}}
\newcommand*{\norm}[1]{\lVert{#1}\rVert}

\providecommand{\sorthelp}[1]{} 

\icmltitlerunning{Listening to the Noise: Blind Denoising with Gibbs Diffusion}

\begin{document}

\twocolumn[
\icmltitle{Listening to the Noise: Blind Denoising with Gibbs Diffusion}



\icmlsetsymbol{equal}{*}

\begin{icmlauthorlist}
\icmlauthor{David Heurtel-Depeiges}{x}
\icmlauthor{Charles C. Margossian}{ccm}
\icmlauthor{Ruben Ohana}{ccm}
\icmlauthor{Bruno Régaldo-Saint Blancard}{ccm}
\end{icmlauthorlist}

\icmlaffiliation{ccm}{Flatiron Institute}
\icmlaffiliation{x}{Ecole Polytechnique, Institut Polytechnique de Paris}

\icmlcorrespondingauthor{David Heurtel-Depeiges}{david.heurtel-depeiges@polytechnique.edu}
\icmlcorrespondingauthor{Bruno Régaldo-Saint Blancard}{bregaldo@flatironinstitute.org}

\icmlkeywords{Machine Learning, ICML}

\vskip 0.3in
]



\printAffiliationsAndNotice{}  

\begin{abstract}

In recent years, denoising problems have become intertwined with the development of deep generative models.
In particular, diffusion models are trained like denoisers, and the distribution they model coincide with denoising priors in the Bayesian picture.
However, denoising through diffusion-based posterior sampling requires the noise level and covariance to be known, preventing \textit{blind denoising}. We overcome this limitation by introducing Gibbs Diffusion (GDiff), a general methodology addressing posterior sampling of both the signal and the noise parameters. 
Assuming arbitrary parametric Gaussian noise, we develop a Gibbs algorithm that alternates sampling steps from a conditional diffusion model trained to map the signal prior to the family of noise distributions, and a Monte Carlo sampler to infer the noise parameters.
Our theoretical analysis highlights potential pitfalls, guides diagnostic usage, and quantifies errors in the Gibbs stationary distribution caused by the diffusion model.
We showcase our method for 1) blind denoising of natural images involving colored noises with unknown amplitude and spectral index, and 2)~a cosmology problem, namely the analysis of cosmic microwave background data, where Bayesian inference of ``noise'' parameters means constraining models of the evolution of the Universe.

\end{abstract}

\section{Introduction}

Denoising is an old problem in signal processing, which has experienced significant advancements in the last decade, propelled by the advent of deep learning~\citep[][]{Tian2020, Elad2023}. Convolutional neural networks have led to the emergence of state-of-the-art image denoisers~\citep[e.g.,][]{Zhang2017, Zhang2022}, by learning complex prior distributions of target signals implicitly. Denoisers were also found to be powerful tools for generative modeling, as strikingly demonstrated by diffusion models~\citep{ho2020denoising, song2021scorebased, Saharia2022, Rombach2022}. Considerable effort has been devoted to recovering highly accurate, noise-free versions of contaminated signals, often at the expense of fully addressing the noise's complexity. In numerous practical scenarios, both in industry and scientific research, accurately characterizing the noise itself is of paramount importance (e.g., medical imaging, astronomy, speech recognition, financial market analysis). This paper addresses the challenge of \textit{blind denoising}, with the objective of simultaneously recovering both the signal and the noise characteristics.

\begin{figure}
    \centering
    \begin{tikzpicture}[node distance=2cm, semithick, rounded rectangle/.style={draw, rounded corners}]

  \node[draw, circle, minimum size=8mm, inner sep=1pt] (phi) at (0.5,-0.3) {$\ve{\phi}$};
  \node[draw, circle, minimum size=8mm, align=center, outer sep=1pt, inner sep=1pt] (eps) at (2,-0.3) {$\ve{\varepsilon}$};
  \node[draw, circle, align=center,minimum size=8mm, inner sep=1pt] (x) at (2,-1.7) {$\ve{x}$};
  \node[draw, circle, fill=gray!20,minimum size=8mm, align=center, inner sep=1pt] (y) at (3,-1) {$\ve{y}$};

  \draw[->] (phi) -- (eps);
  \draw[->] (eps) -- (y);
  \draw[->] (x) -- (y);

\end{tikzpicture}
    \caption{Graphical model: we observe $\ve{y} = \ve{x} + \ve{\varepsilon}$ and aim to infer both $\ve{x}$ and $\ve{\phi}$ in a Bayesian framework.}
    \label{fig:graphical_model}
\end{figure}
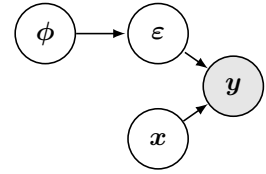

We formalize the problem as follows. We observe a signal $\ve{y} \in \mathbb{R}^d$ that is an additive mixture of an arbitrary signal $\ve{x}$ and a Gaussian signal $\ve{\varepsilon}$ with 
covariance $\ve{\Sigma}_\ve{\phi} \in \mathbb{R}^{d\times d}$:
\begin{align}
    \label{eq:denoising_pb}
    \ve{y} = \ve{x} + \ve{\varepsilon},~\text{with}~\ve{\varepsilon} \sim \mathcal{N}(\ve{0}, \ve{\Sigma}_\ve{\phi}),
\end{align}
where $\ve{\phi} \in \mathbb{R}^K$ is an unknown vector of parameters (see Fig.~\ref{fig:graphical_model} for a graphical model). 
The functional form of $\ve{\Sigma}_\ve{\phi}$ can be arbitrary, although computational constraints would typically require that the number of parameters $K$ remains reasonable. For example, the vector $\ve{\phi}$ would typically include a parameter $\sigma > 0$ controlling the overall noise amplitude, but could also encompass parameters describing the local variations or spectral properties of the noise (cf Sect.~\ref{sec:applications}).
We frame the problem in a Bayesian picture, where prior information on $\ve{x}$ and $\ve{\phi}$ is given, representing some pre-existing knowledge or assumptions on the target data. We will assume that the prior information over $\ve{\phi}$ takes the form of an analytical prior distribution $p(\ve{\phi})$. Crucially, although the prior distribution $p(\ve{x})$ on the signal $\ve{x}$ is typically analytically intractable, we assume access to a set of examples $\ve{x}_1, \dots, \ve{x}_N$ drawn from $p(\ve{x})$. In this Bayesian context, solving \textit{blind denoising} amounts to sampling the posterior distribution $p(\ve{x}, \ve{\phi}\,|\,\ve{y})$.

\paragraph{Contributions.} 1) We introduce Gibbs Diffusion (GDiff), a novel approach for blind denoising (Sect.~\ref{sec:method}). The method combines diffusion models with a Gibbs sampler, tackling simultaneously the challenges of modeling the prior distribution $p(\ve{x})$ based on the samples $\ve{x}_1, \dots, \ve{x}_N$, and sampling of the posterior $p(\ve{x}, \ve{\phi}\,|\,\ve{y})$. 2) We establish conditions for the stationary distribution's existence and quantify inference error propagation. 3) We showcase our method in two areas (Sect.~\ref{sec:applications}). First, we tackle blind denoising of natural images contaminated by arbitrary colored noise, outperforming standard baselines. Second, we demonstrate the interest of our method for cosmology, where Bayesian inference of ``noise'' parameters corresponds to constraining cosmological parameters of models describing the Universe's evolution. 4)~We provide our code on \href{https://github.com/rubenohana/Gibbs-Diffusion}{GitHub}.\footnote{\url{https://github.com/rubenohana/Gibbs-Diffusion}}

\paragraph{Related work.} Blind denoising has garnered substantial attention in the literature, especially in image processing, where denoisers based on convolutional neural networks have become the standard~\citep[e.g.,][]{Zhang2017, Batson2019, ElHelou2020}. However, these techniques predominantly focus on white Gaussian noises and typically aim for point estimates such as the minimum mean square error or maximum a posteriori estimates, often neglecting the quantification of uncertainties in both the signal and the noise parameters. With the advent of diffusion models, innovative methods for removing structured noise have been developed~\citep{Stevens2023}, and posterior sampling has become achievable~\citep[e.g.,][]{Heurtel2023, xie2023diffusion}. However, to the best of our knowledge, no existing work has yet fully addressed blind denoising using diffusion-based priors.

Denoising problems can be seen as a subset of the broader category of linear inverse problems, for which observations read $\ve{y} = \ve{A}\ve{x} + \ve{\varepsilon}$ with $\ve{A}$ a general linear operator. Diffusion models have proven to be of great interest for these problems, generating a wealth of literature adjacent to our problem. In various techniques, diffusion-based priors have been integrated as plug-and-play models~\citep[see e.g.,][]{Kadkhodaie2021, meng2022diffusion, Kawar2022, chung2023diffusion, Song2023, zhu2023denoising, rout2023solving}, relying on the decomposition of the conditional score $\nabla_{\ve{x}_t}\log p_t(\ve{x}_t\,\vert\,\ve{y})$. These methods typically involve approximations of the guiding term $\nabla_{\ve{x}_t}\log p_t(\ve{y}\,\vert\,\ve{x}_t)$. In Appendix \ref{app:other_methods}, we discuss limitations of some of these methods for our problem. Still employing the diffusion model as a plug-and-play prior but avoiding such approximations, \citet{Cardoso2023} introduced a sequential Monte Carlo method for Bayesian inverse problems. Interestingly, \citet{GibbsDDRM} developed a Gibbs sampler to address the case of a linear operator $\ve{A}$ depending on unknown parameters $\ve{\phi}$. While the methodology bears similarities with the approach taken in this paper, it does not seem easily adaptable to the blind denoising problem we address in this work. Finally, let us mention \citet{laroche2024fastem}, which addresses blind deconvolution problems by estimating the maximum a posteriori using an Expectation-Minimization algorithm combined with a guidance-based sampling method.

\section{Method}
\label{sec:method}

Our method builds on the ability of diffusion models to perform posterior sampling in denoising contexts. We elaborate on this aspect in Sect.~\ref{sec:method_ss}, a topic we found to be not directly addressed in the existing denoising literature. Then, in Sect.~\ref{sec:method_gibbs}, we introduce GDiff, a method that addresses posterior sampling of $p(\ve{x}, \ve{\phi}\,|\,\ve{y})$. Finally, we study theoretical properties of GDiff in Sect.~\ref{sec:gibbs_properties}.

\subsection{Diffusion Models as Denoising Posterior Samplers}
\label{sec:method_ss}

\begin{figure}
\centering
\begin{tikzpicture}[scale = 1]
\pgfmathsetlengthmacro\MajorTickLength{
      \pgfkeysvalueof{/pgfplots/major tick length} * 0.5}
    
\node[ellipse, draw = black, text = black, fill = white, minimum width = 1.2cm, minimum height = 2cm] (e) at (0,0) {$p_0(\ve{z}_0)$};
\node at (0, 1.3) {$t=0$};

\node[ellipse, draw = black, text = black, fill = white, minimum width = 1.2cm, minimum height = 2cm] (e) at (4.0,0) {$p_1(\ve{z_1})$};
\node at (4.0, 1.3) {$t=1$};

\draw [<->, densely dotted, line width = 1] (3.4, 0.2) arc [start angle = 30, end angle = 150, x radius = 1.6cm, y radius = 0.5cm];

\path[draw = blue1, fill = white, line width = 1] (2.4, 0.34) -- (2.5, 0.44) -- (2.4, 0.54) -- (2.3, 0.44) -- cycle;

\draw [->, blue1, line width = 1] (2.4, -0.3) -- (2.4, 0.2);

\node at (2.2, -0.6) {\small $\tilde{\ve{y}} \sim p_{t^*} ( \ve{z}_{t^*} | \ve{\phi})$};

\end{tikzpicture}
\caption{We train diffusion models to define a stochastic linear interpolation between the signal prior distribution $p(\ve{x})$ and the noise distribution $p(\ve{\ve{\varepsilon'}}) \sim \mathcal{N}(\ve{0}, \ve{\Sigma}_{\ve{\varphi}})$.}
\label{fig:ddpm_posterior_sampling}
\end{figure}
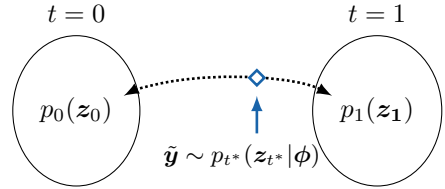

In a non-blind denoising setting, noise parameters $\ve{\phi}$ are known. In the Bayesian picture, solving a denoising problem means sampling the posterior distribution $p(\ve{x}\,|\,\ve{y}, \ve{\phi})$ for a given prior $p(\ve{x})$. We show here, that a diffusion model can be naturally trained to both define a prior $p(\ve{x})$ and yield an efficient means to sample the posterior distribution $p(\ve{x}\,|\,\ve{y}, \ve{\phi})$. This idea was leveraged in~\citet{Heurtel2023} for a scientific application in cosmology. This was also the concern of \citet{xie2023diffusion}.

Diffusion models are a class of deep generative models that are trained to reverse a process consisting in gradually adding noise to the target data. These models are well described using the formalism of stochastic differential equations~\citep[SDE,][]{song2021scorebased}.

The noising process is called the forward SDE, and we only consider here noising Itô processes $(\ve{z}_t)_{t\in [0, 1]}$ defined by:
\begin{equation}
\label{eq:forward_sde}
\begin{cases}
    \drm\ve{z}_t = -f(t)\ve{z}_t\,\drm t + g(t)\ve{\Sigma}_\ve{\varphi}^{\frac12}\,\drm \ve{w}_t,\\
    \ve{z}_0 = \ve{x},
\end{cases}
\end{equation}
where $f, g : [0, 1] \rightarrow \mathbb{R}_+$ are two continuous functions, $(\ve{w})_{t\in [0, 1]}$ is a standard $d$-dimensional Wiener process, and $\ve{\Sigma}_\ve{\varphi}$ corresponds to a normalized version of the covariance matrix $\ve{\Sigma}_\ve{\phi}$ introduced in Eq.~\eqref{eq:denoising_pb} (e.g., $\ve{\Sigma}_\ve{\varphi} = \ve{\Sigma}_\ve{\phi} / \norm{\ve{\Sigma}_\ve{\phi}}$). Within the noise parameters $\ve{\phi}$, we make a distinction between the noise amplitude $\sigma$ (i.e. typically $\norm{\ve{\Sigma}_\ve{\phi}}^{1/2}$), and the rest of the parameters represented as $\ve{\varphi}$. These play distinct roles in the noising process: $\ve{\varphi}$ modulates the structure of the noise while $\sigma$ is muddled with time $t$ (see App.~\ref{app:proof_lemma}). In this setting, the following proposition applies.

\begin{proposition}
\label{lemma:fwd}
For a forward SDE $(\ve{z})_{t\in [0, 1]}$ defined by Eq.~\eqref{eq:forward_sde}, for all $t \in [0, 1]$, $\ve{z}_t$ reads
\begin{equation}
	\ve{z}_t = a(t)\ve{x} + b(t)\ve{\varepsilon^\prime},\hfill \text{with}~ \varepsilon^\prime \sim \mathcal{N}(0, \ve{\Sigma}_\ve{\varphi}),
\end{equation}
with $a : [0, 1] \rightarrow \mathbb{R}_+$ a decreasing function with $a(0) = 1$, and $b : [0, 1] \rightarrow \mathbb{R}_+$ an increasing function with $b(0) = 0$.\footnote{See App.~\ref{app:proof_lemma} for expressions of $a$ and $b$ as functions of $f$ and $g$.}

Moreoever, for $f$ and $g$ suited so that $\sigma\leq b(1)/a(1)$, there exists $t^* \in [0, 1]$ such that
\begin{equation}
	\tilde{\ve{y}} \defeq a(t^*)\ve{y} \laweq \ve{z}_{t^*}.
\end{equation}
\end{proposition}

\textit{Proof:} See App.~\ref{app:proof_lemma}.

\textit{Remark:} Forward SDEs of denoising probabilistic diffusion models~\citep[DDPM,][]{Sohl2015, ho2020denoising} are particular cases of Eq.~\eqref{eq:forward_sde} where $f(t) = \beta(t) / 2$ and $g(t) = \sqrt{\beta(t)}$ for an increasing function $\beta : [0, 1] \rightarrow \mathbb{R}_+$.

Denoting by $p_t(\ve{z}_t)$ the distribution of $\ve{z}_t$, Prop.~\eqref{lemma:fwd} presents the forward process $(\ve{z})_{t\in [0, 1]}$ as a stochastic linear interpolant between $p_0(\ve{z}_0) \sim p(\ve{x})$ and $p_1(\ve{z}_1)$~\citep[for a broader perspective, see][]{Albergo2023}. In particular, provided that $\sigma \leq b(1)/a(1)$, our observation $\ve{y}$ can be viewed as a rescaled realization of $\ve{z}_{t^*}$ with $t^*\in [0, 1]$~(see Fig.~\ref{fig:ddpm_posterior_sampling} for an illustration).

The existence of a reverse process associated with the forward SDE is proved in \citet{anderson1982reverse}.  For the forward SDE~\eqref{eq:forward_sde}, it takes the form of a process $(\bar{\ve{z}}_t)_{t\in [0, 1]}$ defined by:
\begin{align}
\label{eq:backward_sde}
\begin{cases}
    \drm\bar{\ve{z}}_t = \left[-f(t)\bar{\ve{z}}_t-g(t)^2\ve{\Sigma}_{\ve{\varphi}}\nabla_{\bar{\ve{z}}_t}\log p_t(\bar{\ve{z}}_t)\right]\,\drm t \\
    \hspace{1cm}+ g(t)\ve{\Sigma}_\ve{\varphi}^{\frac12} \,\drm \overline{\ve{w}}_t, \\
    \bar{\ve{z}}_1 \laweq \ve{z}_1
\end{cases}
\end{align}
where time flows backward and $\overline{\ve{w}}_t$ denotes a standard backward-time Wiener process. According to \citet{anderson1982reverse}, $(\bar{\ve{z}}_t)_{t\in [0, 1]}$ and $(\ve{z}_t)_{t\in [0, 1]}$ are equal in law. This implies that the marginal distributions of $(\bar{\ve{z}}_t)$ and $(\ve{z}_t)$ coincide for all $t$, but also that the joint distributions $p(\ve{z}_{t_1}, \ve{z}_{t_2})$ and $p(\bar{\ve{z}}_{t_1}, \bar{\ve{z}}_{t_2})$ coincide for all pair $(t_1, t_2) \in [0,1]^2$. Thanks to this latter fact, sampling $p(\ve{z}_0\,|\ve{z}_{t^*})$ can be achieved by solving the reverse SDE~\eqref{eq:backward_sde} starting from time $t=t^*$ and initialization $\bar{\ve{y}}$. This sampling procedure naturally yields an (approximate) sample of $p(\ve{x}\,|\,\ve{y}, \ve{\phi})$.


\begin{algorithm}[tb]
   \caption{GDiff: Gibbs Diffusion for Blind Denoising}
   \label{alg:gibbs_sampler}
\begin{algorithmic}
   \STATE {\bfseries Input:} observation $\ve{y}$, $p_{\rm init}(\ve{\phi})$, $M$
   \STATE Initialize $\ve{\phi}_0 \sim p_{\rm init}(\ve{\phi})$.
   \STATE Initialize $\ve{x}_0 \sim q(\ve{x}\,|\,\ve{y}, \ve{\phi}_0)$\hfill \text{(Diffusion step)}
   \FOR{$k=1$ {\bfseries to} $M$}
   \STATE $\ve{\varepsilon}_{k-1} \leftarrow \ve{y} - \ve{x}_{k-1}$
   \STATE  $\ve{\phi}_k \sim q(\ve{\phi}\,|\,\ve{\varepsilon}_{k-1})$ \hfill \text{(HMC step)}
   \STATE $\ve{x}_k \sim  q(\ve{x}\,|\,\ve{y}, \ve{\phi}_k)$\hfill \text{(Diffusion step)}
   \ENDFOR 
   \STATE \textbf{Output:} samples $(\ve{x}_k, \ve{\phi}_k)_{1 \leq k \leq M}$. 
\end{algorithmic}
\end{algorithm}

\subsection{Blind Denoising with Gibbs Diffusion}
\label{sec:method_gibbs}

We now address the problem of solving blind denoising, where the noise parameters $\ve{\phi}$ are to be inferred. In our Bayesian picture, the goal is to sample the joint posterior distribution $p(\ve{x}, \ve{\phi}\,|\,\ve{y})$, for given priors $p(\ve{x})$ and $p(\ve{\phi})$.

\paragraph{Gibbs Sampling.} GDiff takes the form of a Gibbs algorithm that iteratively constructs a Markov chain $(\ve{x}_k, \ve{\phi}_k)_{0 \leq k \leq M}$ by alternating the sampling between the conditional distributions $p(\ve{x}\,|\,\ve{y}, \ve{\phi})$ and $p(\ve{\phi}\,|\,\ve{y}, \ve{x})$. For ideal sampling, after an initial warm-up phase allowing the system to reach its stationary regime, the iterations of the chains produce samples from the joint distribution $p(\ve{x}, \ve{\phi}\,|\,\ve{y})$. 

\begin{enumerate}
\item \textbf{Sampling of $p(\ve{x}\,|\,\ve{y}, \ve{\phi})$}. Sect.~\ref{sec:method_ss} has shown that a diffusion model can be trained to naturally address posterior sampling of $p(\ve{x}\,|\,\ve{y}, \ve{\phi})$ provided that the forward process meets certain conditions. In practice, our forward process must take the form of Eq.~\eqref{eq:forward_sde} and $f$ and $g$ have to be such that $\sigma\leq b(1)/ a(1)$, which is always easily achievable (cf Sect.~\ref{sec:applications}). Moreover, since $\ve{\phi}$ is a priori unknown, we train a conditional diffusion model where $\ve{\phi}$ is given an an input to the score network, using a range of values consistent with the prior distribution $p(\ve{\phi})$.

\item \textbf{Sampling of $p(\ve{\phi}\,|\,\ve{y}, \ve{x})$}. If $\ve{y}$ and $\ve{x}$ are known, then $\ve{\varepsilon} = \ve{y} - \ve{x}$ is known, so that sampling $p(\ve{\phi}\,|\,\ve{y}, \ve{x})$ is equivalent to sampling $p(\ve{\phi}\,|\,\ve{\varepsilon})$. We address the sampling of $p(\ve{\phi}\,|\,\ve{\varepsilon})$ using a Hamiltonian Monte Carlo (HMC) sampler~\citep[see][for a review]{Neal2010, Betancourt2018}. Provided we can efficiently evaluate and differentiate the (unnormalized) target distribution $p(\ve{\phi}\,|\,\ve{\varepsilon})$  with respect to $\ve{\phi}$, a HMC sampler constructs a Markov chain that yields (weakly) correlated samples of the target distribution once the stationary regime is reached. In our case, the log posterior distribution $\log p(\ve{\phi}\,|\,\ve{\varepsilon}) = \log\left[p(\ve{\varepsilon}\,|\,\ve{\phi})\right] + p(\ve{\phi}) + C_{\ve{\varepsilon}}$ requires the ability to evaluate and differentiate efficiently the prior distribution (assumed to be analytically tractable) and the log likelihood $\log p(\ve{\varepsilon}\,|\,\ve{\phi})$, which reads:
\begin{align}
\label{eq:log_likelihood_hmc}
    \log\left[p(\ve{\varepsilon}\,|\,\ve{\phi})\right] = -\frac{1}{2}\log\left[\det(\ve{\Sigma}_\ve{\phi})\right]
    -\frac{1}{2}\ve{\varepsilon}^T\ve{\Sigma}^{-1}_\ve{\phi} \ve{\varepsilon} + C,
\end{align}
due to the Gaussianity of $\varepsilon$. This can be easily achieved for the applications considered in Sect.~\ref{sec:applications}. It is important to note, however, that our Gibbs algorithm does not strictly require HMC for this step. Alternative sampling strategies (e.g., neural estimation with normalizing flows) could also be employed and may be more appropriate for other types of applications.
\end{enumerate}

\paragraph{Algorithm and practical details.} 
Algo~\ref{alg:gibbs_sampler} describes our algorithm.
It requires specifying the number of Gibbs iterations $M$, which includes both the iterations needed for the warm-up phase and the desired number of output samples from the target distribution.  Additionally, an initialization strategy for the chains must be defined, represented by the distribution $p_{\rm init}$. In theory, any strategy of initialization of $\ve{\phi}$ is viable, as the stationary distribution is independent from $p_{\rm init}$. A natural choice for this is the prior distribution $p(\ve{\phi})$. However, in practice, to aid in the convergence of the Gibbs sampler, initial guesses informed by simple heuristics or auxiliary inference strategies might be advantageous.

\subsection{Properties of the Gibbs Sampler}
\label{sec:gibbs_properties}

Under certain regularity conditions, the ordinary Gibbs sampler converges to the posterior distribution \citep{Geman1984}. 
However, in our setting, as we do not draw samples from the exact conditionals of $\ve{x}$ and $\phi$, additional care must be taken to understand our algorithm's behavior.

Let $T$ be the \textit{transition kernel} of the Gibbs sampler, that is the probability of evolving from a state $(\phi_{k - 1}, \ve{x}_{k - 1})$ to a state $(\phi_k, \ve{x}_{k})$.
Then the stationary distribution $\pi$ is the distribution left invariant by $T$, meaning that if $(\phi_{k - 1}, \ve{x}_{k - 1}) \sim \pi$ and we take one step with the Gibbs sampler, then it is also the case that $(\phi_{k}, \ve{x}_{k}) \sim \pi$.
Now, $T$ is a composition of two transition kernels: $T_\ve{\phi}$, the HMC step, and $T_\ve{x}$, the diffusion model step.
\begin{proposition} \label{prop:invariant}
  The transition kernel $T_\ve{\phi}$ leaves invariant any joint distribution $\pi(\ve{\phi}, \ve{x} \mid \ve{y})$  satisfying $\pi(\ve{\phi} \mid \ve{y}, \ve{x}) = p(\ve{\phi} \mid \ve{y}, \ve{x})$, while $T_\ve{x}$ leaves invariant any joint distribution $\tilde \pi(\ve{\phi}, \ve{x} \mid \ve{y})$ satisfying $\tilde \pi(\ve{x} \mid \ve{y}, \ve{\phi}) = q(\ve{x} \mid \ve{y}, \ve{\phi})$.
\end{proposition}
\textit{Proof.} The proof is detailed in App.~\ref{app:Gibbs}.

$T_\phi$ has the same invariant distribution, whether we use an HMC step or exact sampling from $p(\ve{\phi} \mid \ve{y}, \ve{x})$.
Therefore, the details of $T_\phi$, such as the number of HMC steps we take per iteration and the tuning of HMC itself impact convergence rate but not the stationary distribution.
On the other hand, using a diffusion model rather than sampling from $p(\ve{x} \mid \ve{y}, \phi)$ changes the stationary distribution.

\paragraph{Existence of the stationary distribution.}
In general, there may not exist a joint distribution over $\ve{x}$ and $\phi$, which matches the conditionals $p(\ve{\phi} \mid \ve{y}, \ve{x})$ and $q(\ve{x} \mid \ve{y}, \ve{\phi})$, in which case we say the conditionals are \textit{incompatible}.
Incompatibility can notably arise if $q(\ve{x} \mid \ve{y}, \phi)$ poorly approximates the true conditional $p(\ve{x} \mid \ve{y}, \phi)$ \citep[e.g.,][]{Hobert1998, Liu2012}.
We show in App.~\ref{app:Gibbs} that the main necessary condition for compatibility is,
\begin{equation} \label{eq:compatible}
  \int_\mathcal{X} \frac{q(\ve{x} \mid \ve{y}, \ve{\phi})}{p(\ve{\phi} \mid \ve{y}, \ve{x})} \text d \ve{x} < \infty.
\end{equation}
%
We adapt this result from \citet[Theorem 1]{Arnold2001} and emphasize in our formulation the relationship between the conditionals.
Conceptually, Eq.~\eqref{eq:compatible} tells us that there can be no measurable set in which $\ve{x}$ is likely given $\phi$ but $\phi$ is unlikely given $\ve{x}$.
If the compatibility condition is verified, then under mild conditions the stationary distribution is unique and we say the two conditionals verify \textit{determinacy}; see \citet{Gourieroux1979} and \citet{Liu:2021} for a more detailed discussion.

\paragraph{Error in the Gibbs sampler.} Assume now compatibility and determinacy. 
Denote $\pi(\ve{\phi}, \ve{x} \mid \ve{y})$ the unique stationary distribution. 
How do the approximations in our Gibbs sampler impact our inference on $\ve{\phi}$?
Let $\rho_M(\ve{\phi} \mid \ve{y})$ be the distribution of $\ve{\phi}$ after $M$ sampling iterations and let $D_\text{TV}$ denote the total variation distance.
Then,
\begin{align} 
    &D_\text{TV} [\rho_M(\ve{\phi} \mid \ve{y}), p(\ve{\phi} \mid \ve{y})] \label{eq:distance} \\ &\le D_\text{TV}[\rho_M(\ve{\phi} \mid \ve{y}), \pi(\ve{\phi} \mid \ve{y})] + D_\text{TV}[\pi(\ve{\phi} \mid \ve{y}), p(\ve{\phi} \mid \ve{y})]. \notag
\end{align}
The first term on the right-hand-side of Eq.~\eqref{eq:distance} is due to non-convergence.
The second term is the error at stationarity due to the diffusion model.
We can formally show that, in general, $\pi(\ve{\phi} \mid \ve{y}) \neq p(\ve{\phi} \mid \ve{y})$ and quantify this error.
\begin{theorem} \label{thm:KL}
  Suppose $p(\ve{\phi} \mid \ve{y}, \ve{x})$ and $q(\ve{x} \mid \ve{y}, \ve{\phi})$ are compatible and are the conditionals of the joint $\pi(\ve{\phi}, \ve{x} \mid \ve{y})$.
  Then the Kullback-Leibler divergence is
  \begin{equation*} \label{eq:KL}
      \text{KL}[p(\ve{\phi} \mid \ve{y}) || \pi(\ve{\phi} \mid \ve{y})] = \mathbb E_{p(\ve{\phi} \mid \ve{y})} \log \mathbb E_{p(\ve{x} \mid \ve{y})} \frac{q(\ve{x} \mid \ve{y}, \ve{\phi})}{p(\ve{x} \mid \ve{y}, \ve{\phi})}.
  \end{equation*}
\end{theorem}
\textit{Proof:} The proof is given in App.~\ref{app:Gibbs}.

\cref{thm:KL} upper-bounds Eq.~\eqref{eq:distance} via Pinsker's inequality,
\begin{equation*}
    D_\text{TV}(p(\ve{\phi} \mid \ve{y}), \pi(\ve{\phi} \mid \ve{y})) \le \frac{1}{2} \sqrt{\text{KL}[p(\ve{\phi} \mid \ve{y}) || \pi(\ve{\phi} \mid \ve{y})]}.
\end{equation*}

Without detailed knowledge of $q(\ve{x} \mid \ve{y}, \phi)$ and $p(\ve{x}, \phi \mid \ve{y})$, we cannot say more about the error in our Gibbs sampler using theory alone.
Compatibility between conditionals has been studied in restricted contexts, for instance when the conditionals belong to certain parametric families \citep{Arnold2001, Liu2012}, but this does not describe our setting where the conditionals are intractable.
Motivated by the variational autoencoder, \citet{Liu:2021} recently proposed a loss function (eq. 2 in reference) which, when minimized to 0, enforces compatibility; checking the viability of this strategy in our setting is left as future work. 
Ergodicity and convergence rates for the Gibbs and HMC samplers are discussed in references \citep{Geman1984, Livingstone2019} but typically, these rates cannot be computed explicitly.

\begin{figure*}[t!]
    \centering
    \raisebox{2em}{\includegraphics[width=0.72\hsize]{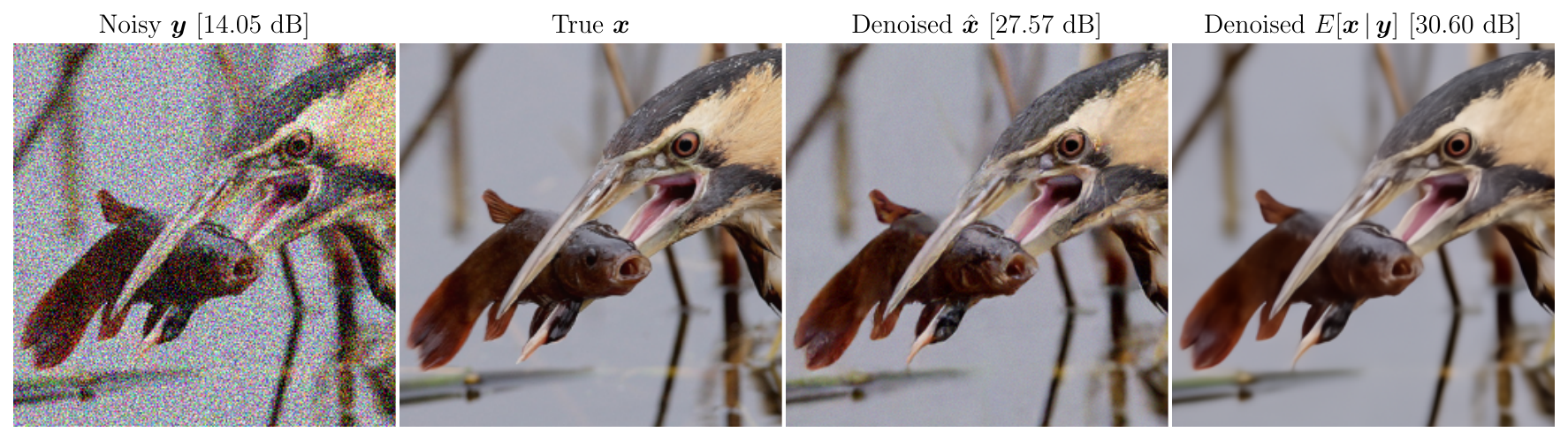}}
    \includegraphics[width=0.25\hsize]{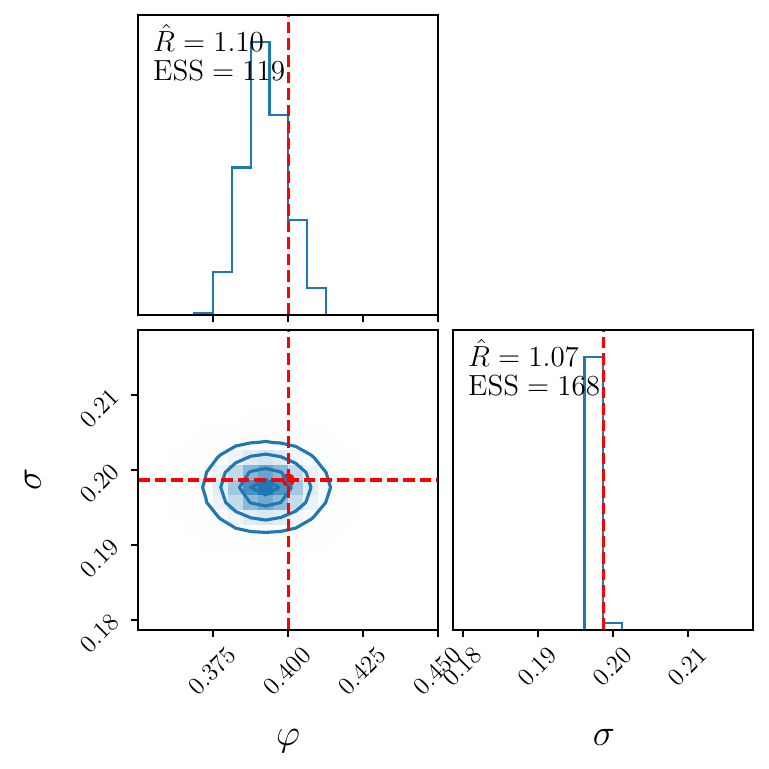}
    \caption{Example of blind denoising with GDiff on an ImageNet sample for $\sigma = 0.2$ and $\varphi = 0.4$. \textit{Left:} Noisy example $\ve{y}$ next to the noise-free image $\ve{x}$, a denoised sample $\hat{\ve{x}}$ and an estimate of the posterior mean $\Eb{\ve{x}\,|\,\ve{y}}$ (with PSNR on top when relevant). \textit{Right:} Inferred posterior distribution over the noise parameters.}
    \label{fig:nat_images_denoising_example}
\end{figure*}

In practice, we recommend empirically checking convergence and the algorithm's calibration.
Convergence diagnostics help us detect if the Gibbs sampler has been run for a sufficient number of iterations and whether a stationary distribution exists.
If poor calibration persists after convergence has been detected, then by Theorem~\ref{thm:KL}, the diffusion model introduces a non-negligible error in our inference.

\section{Applications}
\label{sec:applications}

We apply GDiff in two independent contexts: 1) blind denoising of natural images considering colored noises with unknown amplitude and spectral index, 2) a cosmological problem, namely the analysis of cosmic microwave background (CMB) data, for which the noise is the signal of cosmological interest and its parameters constrain models of the evolution of the Universe.

\paragraph{Common technical details.} In this section, we work with diffusion models relying on the DDPM forward SDE. With the notations of Eq.~\eqref{eq:forward_sde}, we take $f(t) = \beta(t) / 2$ and $g(t) = \sqrt{\beta(t)}$ with $\beta : [0, 1] \rightarrow \mathbb{R}_+$ a linearly increasing function with $\beta(0) = 0$ and $\beta(1)$ calibrated to be higher than the maximum noise level considered in the prior $p(\sigma)$.

Moreover, in both cases, we consider a covariance matrix $\ve{\Sigma}_\ve{\phi}$ that is diagonal in Fourier space. We write ${\ve{\Sigma}_\ve{\phi} = \sigma^2 \ve{F}^T \ve{D}_{\ve{\varphi}} \ve{F}}$, where $\ve{F}$ is the (orthonormal) discrete Fourier transform (DFT) matrix, $\sigma > 0$ controls the noise amplitude and $\ve{D}_{\ve{\varphi}}$ is a diagonal matrix parametrized by $\ve{\varphi} \in \mathbb{R}^{K-1}$. With these notations, $\ve{\phi}$ represents the pair $(\sigma, \ve{\varphi})$. We identify $\ve{D}_{\ve{\varphi}}$ to a power spectrum function $S_\ve{\varphi}$, in the sense that the diagonal coefficients of $\ve{D}_{\ve{\varphi}}$ correspond to the evaluation of $S_{\ve{\varphi}}$ on a discrete set of Fourier modes $\ve{k}$. In this context, the expression of the log likelihood $\log p(\ve{\varepsilon}\,|\,\ve{\phi})$ in Eq.~\eqref{eq:log_likelihood_hmc} takes a simpler form:
\begin{align}
\label{eq:log_likelihood_hmc_simplified}
\log\left[p(\ve{\varepsilon}\,|\,\ve{\phi})\right] = &-\frac{1}{2}\sum_{\ve{k}}\log \sigma^2 S_\ve{\varphi}(\ve{k}) -\frac{1}{2}\sum_{\ve{k}} \frac{\lvert\hat{\ve{\varepsilon}}_\ve{k}\rvert^2}{\sigma^2 S_\ve{\varphi}(\ve{k})} \nonumber \\
&+ C,
\end{align}
where $\hat{\ve{\varepsilon}}$ denotes the DFT of $\ve{\varepsilon}$. Technical details on the HMC algorithm employed in this section are given in App.~\ref{app:hmc_details}. 

\subsection{Denoising for Natural Images}
\label{sec:nat_images}

\paragraph{Data and setting.} We address blind denoising on natural images contaminated by \textit{colored noises}, that it stationary noises with power spectrum following a power law $S_{\varphi}(\ve{k}) = \lvert \ve{k} \rvert^{\varphi}$ for $k \neq 0$ and $S_{\varphi}(\ve{0}) = 1$, with $\varphi \in \mathbb{R}$. We train our diffusion model on the ImageNet data set~\citep{Deng2009, russakovsky2015imagenet}, made of $\sim1.2$~M images, generating noises with parameters drawn from a uniform prior $(\sigma, \varphi) \sim p(\ve{\phi}) = \mathcal{U}([0, 1]\times[-1, 1])$. For the validation of our method, we will also consider a heldout set of ImageNet images, as well as images from the CBSD68 dataset~\citep{Martin2001}. The latter dataset will serve as a means to quantify the transfer properties of our model.

\paragraph{Architecture and training.} Our diffusion model relies on a U-net architecture with attention layers. It is conditioned by both the time $t$ and parameter $\varphi$. This architecture is suited to 256x256 RGB images. Input data is rescaled and cropped accordingly. The diffusion model is trained by minimizing the denoising score matching loss in a discrete time setting with 5,000 time steps. Further technical details are provided in App.~\ref{app:training}.

\begin{table*}

{\scriptsize 
\setlength{\tabcolsep}{2pt} 
\centering
\begin{tabular}{ c c || c c c c || c c c c || c c c c}
\multicolumn{1}{c}{\multirow{3}{*}{Dataset}} & \multicolumn{1}{c||}{\multirow{3}{*}{\parbox{1cm}{\centering Noise Level $\sigma$}}} &  \multicolumn{4}{c||}{$\varphi = -1 \rightarrow$ {\color{betterpink} Pink} noise} & \multicolumn{4}{c||}{$\varphi = 0 \rightarrow $ White noise} & \multicolumn{4}{c}{$\varphi = 1 
 \rightarrow$ {\color{blue} Blue} noise}\\
\cline{3-14}
& & \multirow{2}{*}{BM3D} & \multirow{2}{*}{DnCNN} & \multirow{2}{*}{\parbox{1cm}{\centering \textbf{GDiff} $~~\hat{\ve{x}}~~$}}& \multirow{2}{*}{\parbox{1.3cm}{\centering \textbf{GDiff} $\Eb{\ve{x}\,|\,\ve{y}}$ }} & \multirow{2}{*}{BM3D} & \multirow{2}{*}{DnCNN} & \multirow{2}{*}{\parbox{1cm}{\centering \textbf{GDiff} $~~\hat{\ve{x}}~~$}} & \multirow{2}{*}{\parbox{1.3cm}{\centering \textbf{GDiff} $\Eb{\ve{x}\,|\,\ve{y}}$ }} & \multirow{2}{*}{BM3D} & \multirow{2}{*}{DnCNN} & \multirow{2}{*}{\parbox{1cm}{\centering \textbf{GDiff} $~~\hat{\ve{x}}~~$}} & \multirow{2}{*}{\parbox{1.3cm}{\centering \textbf{GDiff} $\Eb{\ve{x}\,|\,\ve{y}}$ }}\\
&&&&&&&&&&&&& \\
\hline 
\multirow{3}{*}{ImageNet} & 0.06 & \mes{31.0}{0.2} & \mes{(30.2)}{0.2} & \mes{29.3}{0.3} & \textbf{\mes{32.2}{0.3}} & \mes{33.7}{0.3} & \mes{33.4}{0.3} & \mes{31.5}{0.3} & \textbf{\mes{34.4}{0.3}} & \mes{34.7}{0.4} & \mes{(33.8)}{0.4} & \mes{32.3}{0.4} & \textbf{\mes{35.3}{0.4}} \\ 
 & 0.1 & \mes{27.8}{0.2} & \mes{(26.8)}{0.1} & \mes{26.7}{0.2} & \textbf{\mes{29.4}{0.2}} & \mes{31.8}{0.3} & \mes{31.8}{0.4} & \mes{29.7}{0.4} & \textbf{\mes{32.7}{0.4}} & \mes{32.1}{0.4} & \mes{(31.5)}{0.3} & \mes{29.9}{0.4} & \textbf{\mes{32.9}{0.3}} \\ 
 & 0.2 & \mes{23.5}{0.2} & \mes{(21.7)}{0.1} & \mes{23.0}{0.3} & \textbf{\mes{25.7}{0.3}} & \mes{28.1}{0.4} & \mes{28.4}{0.4} & \mes{26.5}{0.4} & \textbf{\mes{29.3}{0.4}} & \mes{29.5}{0.4} & \mes{(28.6)}{0.4} & \mes{27.6}{0.4} & \textbf{\mes{30.5}{0.4}} \\ 
\hline
\multirow{3}{*}{CBSD68} & 0.06 & \mes{31.2}{0.2} & \mes{(30.6)}{0.1} & \mes{29.2}{0.2} & \textbf{\mes{32.2}{0.2}} & \mes{33.8}{0.3} & \mes{34.2}{0.3} & \mes{31.2}{0.3} & \textbf{\mes{34.4}{0.3}} & \mes{35.0}{0.3} & \mes{(34.8)}{0.3} & \mes{32.2}{0.3} & \textbf{\mes{35.5}{0.3}} \\ 
 & 0.1 & \mes{27.9}{0.2} & \mes{(26.9)}{0.1} & \mes{26.2}{0.3} & \textbf{\mes{29.1}{0.3}} & \mes{31.3}{0.3} & \mes{31.7}{0.3} & \mes{28.6}{0.3} & \textbf{\mes{31.8}{0.3}} & \mes{33.0}{0.3} & \mes{(32.7)}{0.3} & \mes{30.6}{0.4} & \textbf{\mes{33.8}{0.4}} \\ 
 & 0.2 & \mes{23.5}{0.2} & \mes{(21.7)}{0.1} & \mes{23.0}{0.3} & \textbf{\mes{25.6}{0.2}} & \mes{27.8}{0.3} & \mes{28.2}{0.3} & \mes{25.4}{0.3} & \textbf{\mes{28.5}{0.3}} & \mes{29.6}{0.3} & \mes{(28.9)}{0.2} & \mes{27.4}{0.3} & \textbf{\mes{30.6}{0.3}} \\ 
\end{tabular}
}
\caption{Denoising performance in terms of PSNR ($\uparrow$, in dB) for GDiff (blind) and baselines BM3D (non-blind) and DnCNN (blind). We report mean PSNR and standard error computed on batches of 50 images. For GDiff, we provide performance for both posterior samples $\ve{x}$ and estimates of the posterior mean $\Eb{\ve{x}\,|\,\ve{y}}$. We point out that DnCNN was trained with white noises only, hence results obtained for $\varphi \neq 0$ could be sub-optimal.}
\label{table:natimages_denoising_psnr}
\end{table*}

\paragraph{Initialization.} Algo~\ref{alg:gibbs_sampler} requires to provide an initialization distribution $p_{\rm init}(\ve{\phi})$ to initialize the Markov chains. For the initialization of the spectral index $\varphi$, we simply use the prior distribution $\mathcal{U}([-1, 1])$. To initialize $\sigma$, we design a simple heuristic based on a linear regression of $\sigma$ given observations $\ve{y}$ fitted on a heldout set.

\paragraph{Denoising results.} We demonstrate the performance of our GDiff method on a noisy ImageNet test image (Fig.~\ref{fig:nat_images_denoising_example}) characterized by $\sigma = 0.2$ and $\varphi = 0.4$. We run our Gibbs algorithm on $M=40$ iterations for 16 chains in parallel. We reach approximate convergence in less than 20 iterations. For our analysis, we consider only the final 20 samples from each chain. The figure presents, on the left, the original noisy image $\ve{y}$ alongside the true image $\ve{x}$, a denoised sample $\hat{\ve{x}}$, and the estimate of the posterior mean $\Eb{\ve{x}\,|\,\ve{y}}$ obtained by averaging all retained samples\footnote{Note that the posterior mean estimate could have been alternatively computed  using Tweedie's formula~\citep{Robbins1956, Efron2011}.}. On the right, the inferred posterior distribution $q(\ve{\phi}\,|\,\ve{y})$ is shown, demonstrating tight constraints around the true parameters and indicative of effective inference. Notably, the denoised posterior sample $\hat{\ve{x}}$ significantly enhances the image's peak signal-to-noise ratio (PSNR).
It is well known that optimal PSNR is attained for the posterior mean $\Eb{\ve{x}\,|\,\ve{y}}$ in the Bayesian picture. This is illustrated on this figure where the posterior mean estimate improves the PSNR of the sample by $\sim$3~dB. This is an interesting illustration of the antagonism between PSNR optimization and posterior sampling. The posterior sample is by construction a more realistic reconstruction in the light of the prior. Conversely, while the posterior mean yields a higher PSNR, indicating closer proximity to the true image, it may represent a less probable realization within the prior's distribution. We provide in Fig.~\ref{fig:nat_images_denoising_other_examples} additional denoising examples.

We now benchmark our blind denoiser against established methods, specifically focusing on the PSNR. We compare our performance with that of BM3D~\citep{Dabov2007, Makinen2020} and DnCNN~\citep[color-blind model,][]{Zhang2017} and report our results in Table~\ref{table:natimages_denoising_psnr}. We show mean performance on subsets of 50 images taken from the ImageNet validation set and CBSD68. It is important to note that the DnCNN model was initially trained for white noise conditions only, hence comparisons for $\varphi \neq 0$ should be interpreted with caution.  Interestingly, posterior samples yielded by our method perform worse than BM3D and DnCNN for these metrics. On the contrary, the posterior mean estimates systematically outperform both BM3D and DnCNN. This is a clear success of our algorithm, which in a blind setting can outperform 
state-of-the-art denoisers. Moreover, we believe that these results could be further improved with additional finetuning of the diffusion model architecture. Finally, we report in Table~\ref{table:natimages_denoising_ssim_with_errorbar} equivalent results for the structural similarity index measure (SSIM) metric~\citep{Wang2004}. Similarly, for this metric, posterior mean estimates provided by GDiff systematically outperform BM3D and DnCNN.

\begin{figure}[h]
    \centering
    \includegraphics[width=0.8\hsize]{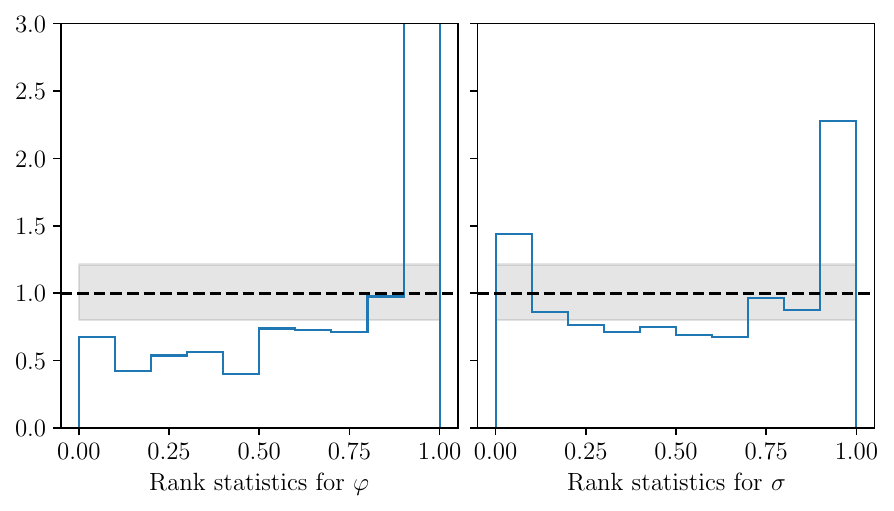}
    \caption{Posterior accuracy diagnostic of GDiff with simulation-based calibration on natural images blind denoising.}
    \label{fig:nat_images_sbc}
\end{figure}

\paragraph{Validation.} We validate our inference pipeline by focusing on the accuracy of the estimation of the noise parameters $\ve{\phi} = (\sigma, \varphi)$. We generate 800 noisy images using the ImageNet validation set, with noise parameters $\ve{\phi}$ sampled from their prior distribution $p(\ve{\phi})$. For each of the noisy image $\ve{y}$, we conduct inferences using 4 parallel chains, each running for $M = 60$ iterations. We discard the initial 30 samples as a warm-up phase. For each noisy observation, we compute the $\hat R$ convergence diagnostic and the effective sample size (ESS)~\citep[see e.g.,][and references therein]{Vehtari2021}. 
Fig.~\ref{fig:nat_images_validation} provides a scatter plot of these diagnostics across the parameter space.
We find that the Markov chains achieve reasonable convergence, as indicated by $\hat{R} \lessapprox 1.1$, and that the ESS per chain is comparable to $M$.
\begin{figure*}[h!]
    \raisebox{1em}{\includegraphics[width=0.58\hsize]{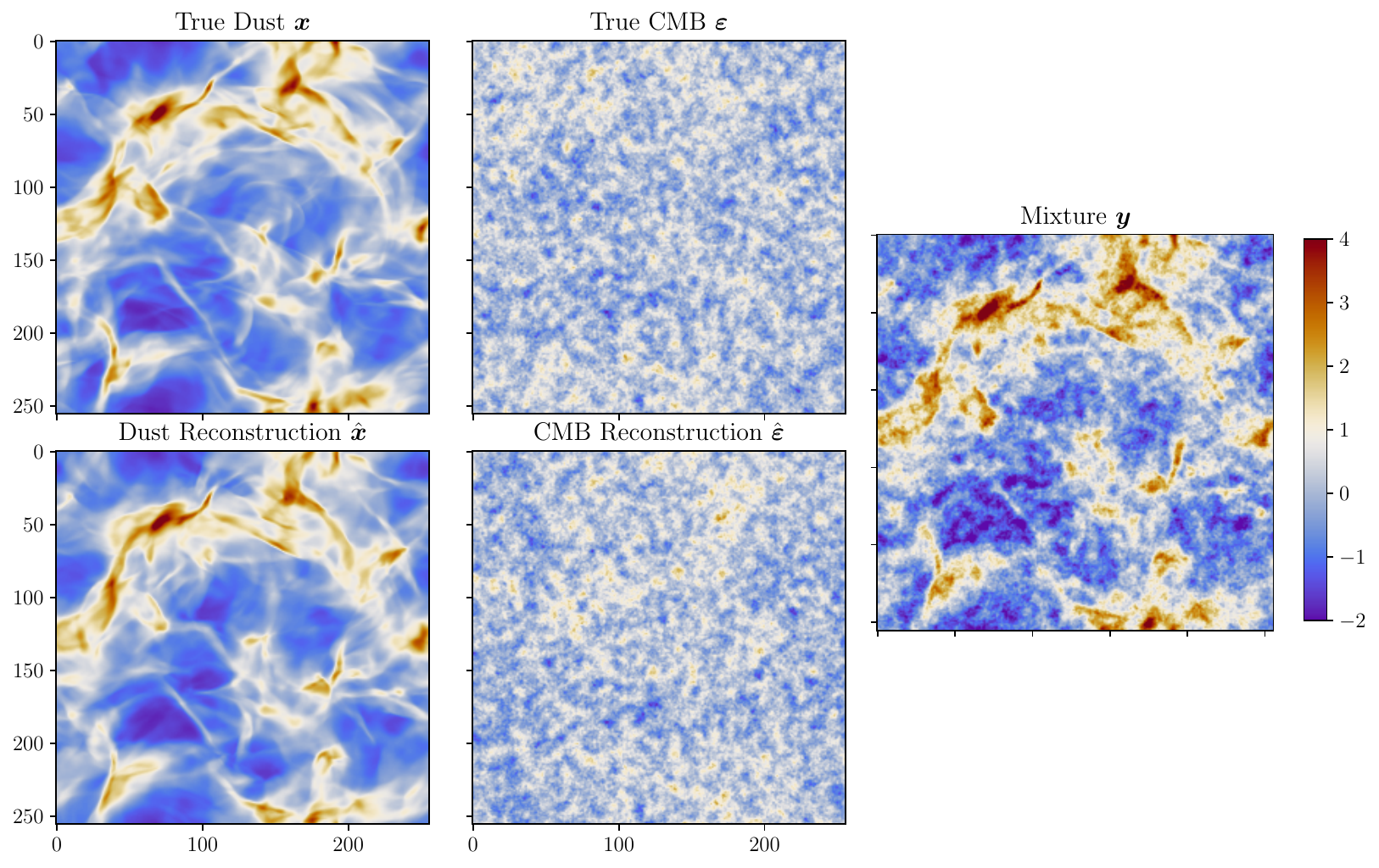}}
    \includegraphics[width=0.38\hsize]{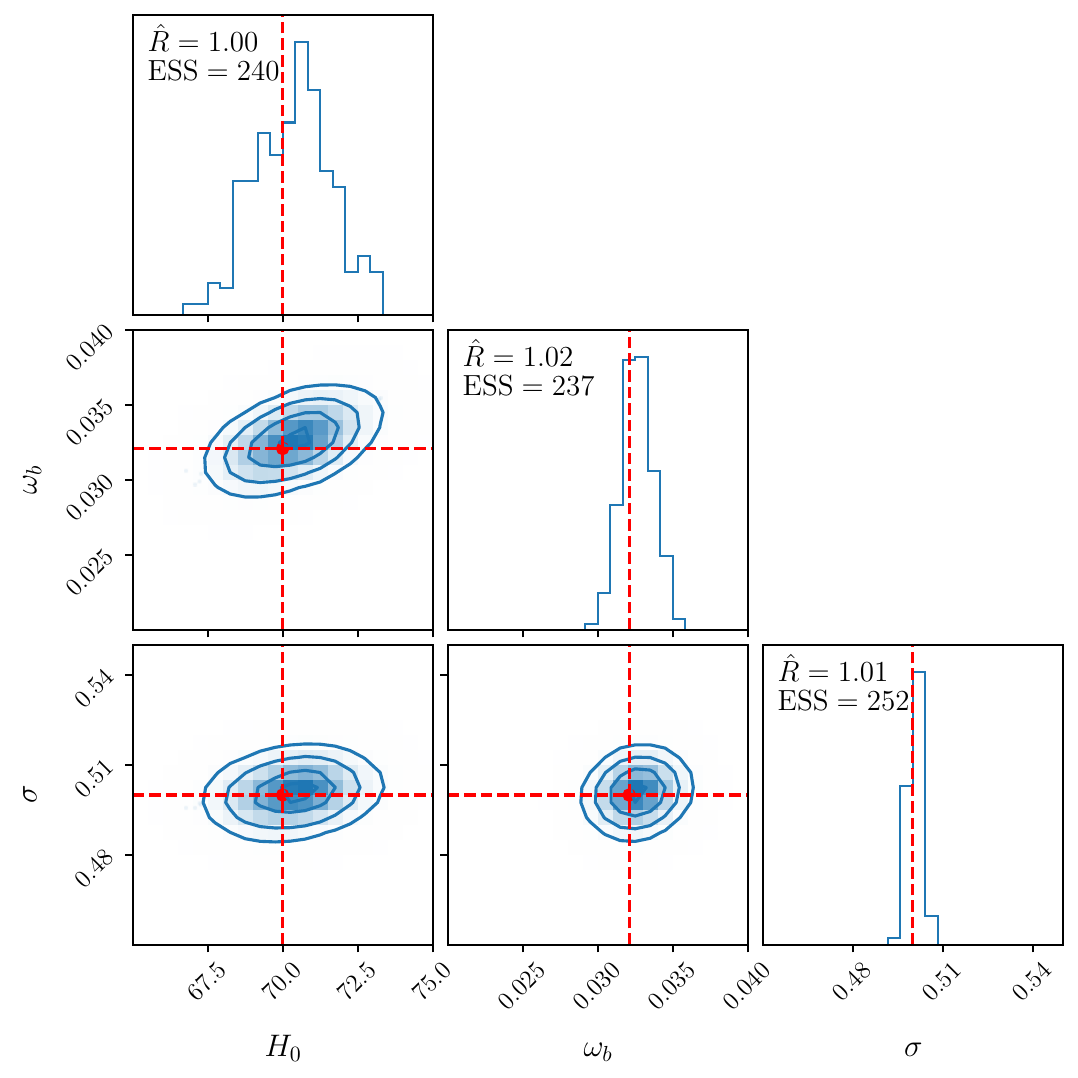}
    \caption{\textit{Left}: Observed maps and maps reconstructed with GDiff. The true dust $\ve{x}$ and CMB $\ve{\varepsilon}$ maps compose the observed mixture $\ve{y}$. We reconstruct  the dust $\hat{\ve{x}}$ and CMB $\hat{\ve{\varepsilon}}$ with our diffusion model. The global unit is arbitrary. \textit{Right:} Inferred cosmological parameters.}
    \label{fig:cosmo_blind_reconstruction}
\end{figure*}

$\hat R$ and ESS tell us how well the Markov chain converges to and explores its stationary distribution, but not how well the stationary distribution approximates the posterior.
To check the latter, we implement the simulation-based calibration diagnostic~\citep[SBC,][]{Talts2018}. This entails computing the empirical distribution of the ranks of the true parameter values within the sampled values. For a well-calibrated inference pipeline, these empirical distributions should follow a uniform distribution. In Fig.~\ref{fig:nat_images_sbc}, we present the resulting normalized rank statistics.  The rank statistics for $\sigma$ align well with a uniform distribution, suggesting reasonable accuracy. However, the rank statistics for $\varphi$ reveal some biases in the predictions. As discussed in Sect.\ref{sec:gibbs_properties}, this discrepancy highlights potential limitations of the diffusion model in accurately modeling the true conditional distribution $p(\ve{x}\,|\,\ve{y}, \ve{\phi})$.

\subsection{Cosmological Inference from CMB Observations}
\label{sec:cosmo_images}

The cosmic microwave background (CMB) is a pivotal cosmological observable for constraining models that describe the Universe's dynamical evolution over its nearly 14 billion-year history~\citep{planck2016-l01}. Yet, CMB observations suffer from the contamination of various astrophysical signals, known as ``foregrounds'', necessitating their removal through precise component separation methods~\citep[e.g.,][]{planck2016-l04}. The quest for primordial $B$-modes in CMB polarization observations~\citep{Kamionkowski2016} underscores the challenge of accurately modeling the thermal emission from interstellar dust grains, or the "dust foreground," which obscures the CMB signal~\citep{pb2015}.

The CMB closely approximates a Gaussian distribution, for which the covariance $\ve{\Sigma}_\ve{\phi}$ relates to cosmological models and $\ve{\phi}$ denotes the target cosmological parameters. Interestingly, component separation for CMB analysis can be viewed as a blind denoising problem of the form of Eq.~\eqref{eq:denoising_pb} where the ``noise'' $\ve{\varepsilon}$ corresponds to the CMB, and the signal $\ve{x}$ represents the foregrounds. Following a rich history of dust foreground modeling efforts~\citep[e.g.,][]{planck2013-p06b, Allys2019, Vansyngel2017, Aylor2021, Regaldo2020, Regaldo2021, Thorne2021, Krachmalnicoff2021, Regaldo2023, harvardworkshop} and going beyond the work of \citet{Heurtel2023}, we focus here on a diffusion-based dust prior trained on simulations and apply GDiff for the Bayesian inference of cosmological parameters $\ve{\phi}$ given a mixture of CMB $\ve{\varepsilon}$ and interstellar dust emission $\ve{x}$.

\paragraph{Data and setting.} We introduce the simulated data in App.~\ref{app:data_cosmo}. The CMB covariance $\ve{\Sigma}_\ve{\phi}$ is parametrized by the CMB amplitude $\sigma$ and cosmological parameters $\ve{\varphi}$. As in \citet{Heurtel2023}, we only consider cosmological parameters $\ve{\varphi} = (H_0, \omega_b)$.  We choose a broad prior $p(\ve{\varphi}) \sim \mathcal{U}([50, 90]\times [0.0075, 0.0567])$. We also consider $\sigma\in[\sigma_{\min},1.2]$ where $\sigma_{\min}$ is taken to be very close to $0$.\footnote{For stability reasons in the diffusion model, $\sigma_{\min}$ must be strictly positive.}

\paragraph{Architecture and training.} The architecture of the diffusion model is a three-level U-net with ResBlocks, trained with a reweighted score matching loss as in Eq.\eqref{eq:reweighted_score_matching_loss}. 
The model is conditioned by both $t$ and parameters $\ve{\varphi}$.
We encountered substantial training instabilities due to the large  ($\sim10^4$) condition number of $\ve{\Sigma}_\ve{\varphi}$. By renormalizing the loss and model outputs, we were able to mitigate these issues. Further technical details are provided in App.~\ref{app:training}.

\paragraph{Initialization.} For this cosmological application, the initialization of the Gibbs sampler plays a more critical role than in Sect.~\ref{sec:nat_images}. Typically, chains initialized far from posterior typical set are likely to converge slowly, if not become stuck in local minima of the posterior distribution. In order to speed-up convergence and improve sampling efficiency, we initialize $\ve{\phi}$ using a CNN trained to estimate the posterior mean $\mathbb{E}[\ve{\phi}\,\vert\,\ve{y}]$~\citep[i.e. \textit{moment network}][]{jeffrey2020solving}. We give further details on this model in App.~\ref{app:hmc_details}.

\begin{figure}[t!]
    \centering
    \includegraphics[width=0.63\hsize]{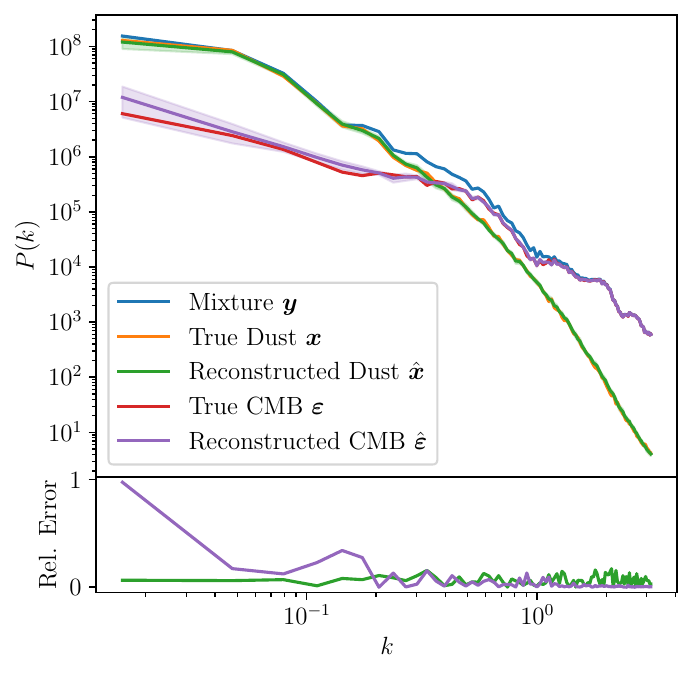}
    \caption{Power spectra of the mixture and the true and reconstructed dust and CMB maps.}
    \label{fig:cosmo_ps}
\end{figure}

\paragraph{Cosmological inference.} We showcase GDiff for this blind component separation context in Fig.~\ref{fig:cosmo_blind_reconstruction} considering a mock mixture $\ve{y}$ with $\sigma = 0.5$ and $(H_0, \omega_b) = (70, 0.032)$. On the left, we show the true maps $\ve{x}$ and $\ve{\varepsilon}$ composing $\ve{y}$, next to a pair of reconstructed samples $\hat{\ve{x}}$ and $\hat{\ve{\varepsilon}}$. On the right, we show a corner plot of the posterior distribution on $\ve{\phi}$ (marginalized over $\ve{x}$). This inferred posterior distribution imposes tight constraints on the cosmological parameters, accurately encompassing the true parameters (marked in red).
We observe that the large scales of the dust map are well reconstructed, as anticipated, due to their lower levels of perturbation. Conversely, at smaller scales, the reconstruction process exhibits more stochastic behavior. In the case of the CMB, the situation is reversed: smaller scales are accurately reconstructed, while the reconstruction of larger scales tends to be more stochastic.
We complement this visual assessment, by a power spectrum analysis, presented in Fig.~\ref{fig:cosmo_ps}. This figure displays the power spectra of both the true and reconstructed dust and CMB maps. The agreement between the true and reconstructed statistics across all scales quantitatively demonstrates the success of our reconstruction process.

\begin{figure}[h!]
    \centering
    \includegraphics[width=\hsize]{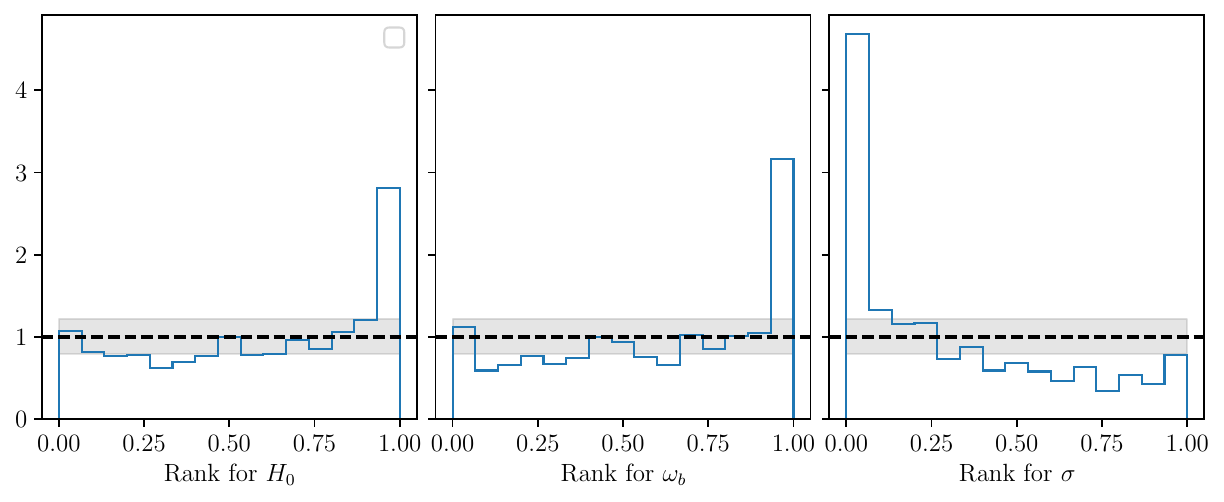}
    \caption{Posterior accuracy diagnostic of GDiff with simulation-based calibration on the cosmological application.}
    \label{fig:cosmo_sbc}
\end{figure}

\paragraph{Validation.} We validate our inference pipeline similarly to Sect.~\ref{sec:nat_images}. The $\hat{R}$ and ESS statistics (see Figs.~\ref{fig:cosmo_images_validation}) for test observations generated across the entire parameter space, indicate that our sampler converges and mixes well. In addition, we conduct SBC for the parameters $\ve{\phi}$ and show in Fig.~\ref{fig:cosmo_sbc} the resulting rank statistics. 
These rank distributions are mostly compatible with the uniform distribution, indicating a well calibrated pipeline in particular for the inference of $H_0$ and $\omega_b$. However, for $\sigma$, the distribution is slanted, indicating bias for some of the chains. Here again, this must be the consequence of errors in the diffusion model, which impact the reconstructed power of the CMB estimates.

\section{Conclusion and Perspectives}
\label{sec:conclusion}

We introduced Gibbs Diffusion (GDiff), a new method to address blind denoising in a Bayesian framework. GDiff takes the form of a Gibbs sampler that alternates sampling of the target signal conditioned on the noise parameters and the observation, and sampling of the noise parameters given the estimated noise. We employed diffusion models for the first step, showing that they can be naturally trained to both address signal prior modeling and posterior sampling of the target signal. For the second step, we opted for a Hamiltonian Monte Carlo sampler, although other alternatives could have been considered. We also derived theoretical properties of the Gibbs sampler, pertaining to existence and invariance properties of the stationary distribution, as well as a quantification of the propagation of errors.

We showcased our method's versatility by applying it to two distinct problems: denoising natural images affected by colored noises with unknown amplitude and spectral index, and performing cosmological inference from simulated cosmic microwave background (CMB) data, which consists of an additive mixture of CMB and a Galactic foreground. Interestingly, while the primary goal in natural image denoising is to recover the noise-free image, the cosmological problem shifts focus to cosmological parameters that characterize the CMB covariance. These parameters assume the role of noise parameters when viewing this task as a blind denoising problem. In both cases, we have shown that our Gibbs sampler converges very well in most situations and constitutes an efficient sampler. Nevertheless, further validation indicated a slight bias due to approximation errors in the diffusion model, suggesting areas for future refinement.

Although diffusion models enable the definition of highly refined prior models, the computational cost of sampling remains significant. Further improvement of our method to accelerate inference would need to take advantage of the recent literature on more efficient ways to address diffusion-based sampling~\citep[e.g.,][]{Rombach2022, Song2023b}. Additionally, our method's applicability is currently limited to scenarios where noise follows a Gaussian distribution. Going beyond this Gaussian assumption while maintaining a rigorous and methodologically sound approach remains an open question, with potentially important applications in signal processing and natural sciences.

\section*{Acknowledgements}

This work benefited from discussions with a number of colleagues whom we wish to acknowledge. In particular, it is a pleasure to thank Yuling Yao for pointing us to a Gibbs sampling approach, as well as Chirag Modi for providing us with a HMC code baseline. We also wish to thank Bob Carpenter and Loucas Pillaud-Vivien for valuable discussions that enriched this work in various ways. Finally, we gratefully acknowledge the Flatiron Institute and the Scientific Computing Core for their support.

\section*{Impact Statement}

This paper presents work whose goal is to advance the fields of Denoising and Machine Learning for Science. There are potential societal consequences of our work, none which we feel must be specifically highlighted here.


\bibliographystyle{bib/icml2024}
\bibliography{bib/references.bib, bib/planck.bib}


\newpage
\appendix
\onecolumn
\counterwithin{figure}{section}

\newpage
\section{Proof of Prop.~\ref{lemma:fwd}}
\label{app:proof_lemma}

Let $f$ and $g$ be two positive real valued functions of time $t$ on the interval $[0,1]$. We consider the following SDE:
\begin{equation}
    \drm\ve{z}_t = f(t) \ve{z}_t\,\drm t + g(t)\,\drm \ve{w}_t,
\end{equation}

\begin{proposition}
For a forward SDE $(\ve{z})_{t\in [0, 1]}$ defined by Eq.~\eqref{eq:forward_sde}, for all $t \in [0, 1]$, $\ve{z}_t$ reads
\begin{equation}
	\ve{z}_t = a(t)\ve{x} + b(t)\ve{\varepsilon^\prime},\hfill \text{with}~ \varepsilon^\prime \sim \mathcal{N}(0, \ve{\Sigma}_\ve{\varphi}),
\end{equation}
with $a : [0, 1] \rightarrow \mathbb{R}_+$ a decreasing function with $a(0) = 1$, and $b : [0, 1] \rightarrow \mathbb{R}_+$ an increasing function with $b(0) = 0$.\footnote{Expressions of $a$ and $b$ as functions of $f$ and $g$ are in App.~\ref{app:proof_lemma}.}

Moreover, for $f$ and $g$ suited so that $\sigma\leq b(1)/a(1)$, there exists $t^* \in [0, 1]$ such that
\begin{equation}
	\tilde{\ve{y}} \defeq a(t^*)\ve{y} \laweq \ve{z}_{t^*}.
\end{equation}
\end{proposition}

\textit{Proof:}
We write here the proof when $\ve{\Sigma}_{\ve{\varphi}}$ is the identity but the proofs holds for any matrix.

If $f$ can be integrated, let $F(t)$ be a primitive of $f$ with value $0$ at time $0$. If $\ve{z}_t$ is an Ito process then so is $\ve{y}_t = \exp(F(t))\ve{z}_t$ and we have:
\begin{align*}
    \drm\ve{y}_t &= \exp(F(t))\drm F(t)\ve{z}_t + \exp(F(t))\drm\ve{z}_t \\
    &= \exp(F(t))f(t)\ve{z}_t,\drm t+ -\exp(F(t))f(t)\ve{z}_t\,\drm t + \exp(F(t))g(t)\,\drm \ve{w}_t\\
    &= \exp(F(t))g(t)\,\drm \ve{w}_t.
\end{align*}
Therefore, integrating this equation from $0$ to $t$ yields:
\begin{equation*}
    \ve{y}_t = \ve{y}_0 + \int_0^t \exp(F(u))g(u)\,\drm \ve{w}_u.
\end{equation*}
And since $\ve{z}_t = \exp(-F(t))\ve{y}_t$, we have:
\begin{equation*}
    \ve{z}_t = \exp(-F(t))\ve{y}_0 + \int_0^t \exp(F(u)-F(t))g(u)\,\drm \ve{w}_u.
\end{equation*}
Using the fact that $F(0) = 0$, we obtain:
\begin{equation}
    \ve{z}_t = \exp(-F(t))\ve{z}_0 + \int_0^t \exp(F(u)-F(t))g(u)\,\drm \ve{w}_u.
\end{equation}
Finally, we have:
\begin{align}
    \int_0^t \exp(F(u)-F(t))g(u)\,\drm \ve{w}_u &\sim \mathcal{N}\left(0, \exp(-2F(t))\int_0^t \exp(2F(u))g(u)^2\,\drm u\right)\\
    \ve{z}_t &\sim \mathcal{N}\left(\exp(-F(t))\ve{z}_0, \exp(-2F(t))\int_0^t \exp(2F(u))g(u)^2\,\drm u\right).
\end{align}

Denoting $a(t)=\exp{-F(t)}$ and $b(t)^2=\exp(-2F(t))\int_0^t \exp(2F(u))g(u)^2\,\drm u$, we have:
\begin{equation}
	\ve{z}_t = a(t)\ve{x} + b(t)\ve{\varepsilon^\prime},\hfill \text{with}~ \varepsilon^\prime \sim \mathcal{N}(0, \ve{\Sigma}_\ve{\varphi}).
\end{equation}

Furthermore, because $f$ is postive, $F$ is increasing and $a$ is a decreasing function of time with $a(0)=1$. If $g$ is not too small\footnote{$\forall t, g(t)\geq b(t) \sqrt{2f(t)}$} then $b$ is increasing. As a consequence, the amount of noise in the mixture $\ve{z}_t$ is $b(t)/a(t)$, itself an increasing function of time and $\forall \sigma\leq b(1)/a(1), \exists t^*\in[0,1], a(t^*)\ve{y}\sim z_{t^*}$.

This is why, for reasonable functions $f$ and $g$, we can identity any mixture of the form $y = z_0 + \sigma \varepsilon$ with a realisation of $\ve{z}_t$ at time $t$ for some $t$ and initial condition $\ve{z}_0$ (this is a stopped forward process, up to a time-dependent rescaling constant). The corresponding time $t$ is often unique for common choices of $f$ and $g$ (both VPSDE, VESDE and all Ornstein-Uhlenbeck like processes used in DM). When tractable, we implement a function $t(\sigma)$ that yields $t^*$ corresponding to $\sigma$ and otherwise, the equation $\sigma=b(t)/a(t)$ is solved via an iterative method.

\section{Additional Details on HMC}
\label{app:hmc_details}

We give details on the practical implementation of the HMC sampler used for the applications of this paper.

\paragraph{Integrator.} During sampling, Hamilton's equations of motion are solved using a leapfrog integrator. Moreover, to improve calibration of the HMC algorithm, for each HMC step, we solve Hamilton's equations over a random number of time steps drawn from a uniform distribution $\mathcal{U}(\{5, \dots, 15\})$. Since both applications of this paper involve compact prior distributions $p(\ve{\phi})$, we also had to implement domain constraints in the integrator. We take into account the prior boundaries by implementing elastic collisions in the leapfrog scheme as described in~\citet{Betancourt2011}.

\paragraph{Initialization Heuristic.} For the cosmological application, initialization of the parameters $\ve{\phi}$ of the Gibbs sampler was found to be more critical than in Sect.\ref{sec:nat_images}. As explained in Sect.\ref{sec:cosmo_images}, we intialize $\ve{\phi}=(\ve{\varphi},\sigma)$ using a CNN with ResBlocks trained to estimate the posterior mean $\mathbb{E}[\ve{\phi}\,\vert\,\ve{y}]$. In practice, we sample $\ve{\varphi},\sigma$ according to their prior distribution and then sample $\ve{y}=\ve{x}+\ve{\varepsilon}$ with $\ve{x}\sim p_\mathrm{data}$ and $\ve{\varepsilon}\sim\mathcal{N}(0,\ve{\Sigma}_\ve{\phi})$. The network is then trained to minimize the MSE loss $\mathbb{E}_\ve{y}[\lvert\lvert\hat{\phi}(\ve{y})-\phi\lvert\lvert^2]$, thus yielding a posterior mean estimator. We found that this moment network was less accurate on the boundary of our (compact) prior. To avoid initializing chains outside the prior, which would get stuck, all posterior mean estimates are projected within the prior domain.

\paragraph{Warm-up.} In the first iteration of the Gibbs sampler (see Algo.~\ref{alg:gibbs_sampler}), we first follow a warm-up phase of $P = 300$ iterations allowing to adapt the step size and estimate a mass matrix. We adapt the step size using a dual averaging procedure~\citep{Nesterov2009} parametrized by a target acceptance rate of 0.65, a regularization scale $\gamma=0.05$, an iteration offset $T_0 = 10$, and a relaxation exponent $\kappa = 0.75$. The inverse of the mass matrix is estimated at iteration $\lfloor 3P/4\rfloor$ using the (unbiased) sample covariance matrix of the warm-up samples after having discarded samples from iterations up to $\lfloor P/4\rfloor$.

\section{Models and Training Details}
\label{app:training}

\subsection{Datasets and Training}
\textbf{Natural images application.} We train our diffusion model on the ImageNet 2012 dataset, which consists of 1,281,167 training images. Images are resized to $3 \times 256 \times 256$ and centered crop. During training, we augment the data with random horizontal flips and rescale the images to the $[0,1]$ range.

We train over 100 epochs on a node of 8 H100 GPUs 80GB with a batch size per GPU of 128 images. Training takes about 41 hours using Data Parallelism. We use the AdamW \citep{loshchilov2017decoupled} optimizer and report good stability of training with respect to learning rate (tested on 0.0001, 0.001, 0.005), with no learning rate schedule. Observing no impact from weight decay, we set its value to 0.

\textbf{Cosmological application.}
\label{app:data_cosmo}
We use the same data as in \citet{Heurtel2023}. We construct simulated dust emission maps in total intensity from a turbulent hydrodynamic simulation of the diffuse interstellar medium taken from the CATS database~\citep{burkhart2020catalogue}. We assume here that dust emission is proportional to the gas density of the simulation. Simulated maps then simply correspond to gas column density maps.
Our dataset consists of 991 dust emission maps of size $256\times256$ (for an example, see Fig.~\ref{fig:cosmo_blind_reconstruction} top left panel). We withheld 10\% of the 991 images to validate the denoising and inference pipeline based on our diffusion model trained on the other 90\%.

If we neglect secondary anisotropies, CMB anisotropies are extremely well described by an isotropic Gaussian random field on the sphere~\citep{planck2016-l01}, entirely characterized by its covariance matrix (or power spectrum). As a consequence, we can obtain CMB maps by sampling Gaussian draws with covariance $\ve{\Sigma}_\ve{\phi}$ parametrized by cosmological parameters $\ve{\varphi}$ (for an example, see Fig.~\ref{fig:cosmo_blind_reconstruction} top middle panel). CMB power spectra are computed using \texttt{CAMB}~\citep{Lewis2000}, and Gaussian draws on the sphere are projected on $256\times 256$ patches using \texttt{pixell}~\footnote{\url{https://github.com/simonsobs/pixell}} with a pixel size of $8^\prime$. In this work, we consider a standard cosmological model. We only vary the cosmological parameters $H_0$ and $\omega_b$ and choose for the remaining ones fiducial values consistent with \textit{Planck} 2018 analysis~\citep{planck2016-l01}.\footnote{We choose $\Omega_K = 0$, $\omega_c = 0.12$, $\tau = 0.0544$, $n_s = 0.9649$, ${\ln(10^{10} A_s) = 3.044}$, $m_\nu = 0.08$.} The resulting simulated maps cover an effective angular surface of approximately $34\times 34\,\mathrm{deg}^2$.

Training examples are generated by combining dust samples and CMB realizations for cosmological parameters randomly drawn from the prior $p(\ve{\phi})$. Since the computation of a single covariance matrix $\ve{\Sigma}_\ve{\phi}$ takes a few seconds, we train neural emulator to approximate $\ve{\phi} \rightarrow \ve{\Sigma}_\ve{\phi}$ as described in the Appendix of \citet{Heurtel2023}. This emulator is differentiable, thus enabling the differentiation of Eq.~\eqref{eq:log_likelihood_hmc_simplified} for HMC.

We train over 100,000 epochs on a single A100 80GB using a batch size of 64 images. Training takes about 96 hours. We use the AdamW optimizer with an inverse square root scheduler with warm-up. We observed that continuing training after the loss had plateaued significantly improved the diffusion model's precision, as evidenced by the simulation-based calibration (SBC) diagnostic.

\subsection{Architectures}\label{app:archi}
\textbf{Natural images application.}
Our diffusion model is a U-Net, that takes images of size 256x256 as an input. It is composed of 5 Downsampling layers, a bottleneck of size 16x16 and 5 Upsampling layers. We add 2 self-Attention layers before the bottleneck and one after the bottleneck. The total number of parameters is $\sim 70,000,000$.

As it is standard in a U-Net architecture, each Upsampling layer is concatenated with the corresponding Downsampling layer. A sinusoidal time embedding is added to each DownConv and UpConv. A 2 layer neural network of inner dimension 100 and SiLU activations are used to embed the exponent $\varphi$ of the colored noise to add this information into the network. We choose a variance preserving DDPM with a discrete time composed of $t_r = 5,000$ time-steps with the schedule $\beta(t) = \frac{0.1}{t_r} + \frac{\beta_{max}-\beta_{min}}{t_r}.\, t$, with $\beta_{min} = 0.1$ and $\beta_{max} = 20$.

\textbf{Cosmological application.} 
For the cosmological application, our score network is a UNet with ResBlocks and a bottleneck of size $32\times 32$ or $16\times 16$. Each ResBlock has three convolution, with GroupNorm as normalization, SiLU for activation and rescaled skip-connection. Long skip connections are concatenated, contrary to internal skip connections who are added. We do not use attention, due to the small size of the dataset. Since both our data and noise have periodic boundary conditions, we use `circular' padding in the convolutions. As in \citep{ho2020denoising,song2021scorebased}, time is also seen as an input. We use the continuous time framework from \cite{song2021scorebased} with Fourier embedding.

In order to manage a parameterized family of SDE (Eq.~\eqref{eq:forward_sde}) we train a diffusion model to reverse each SDE independently by also providing $\varphi$ as an argument to the model. The dependency on $\varphi$ is managed similarly to that of the time $t$ with the exception that the embedding is a linear layer with activation function of the parameters $\varphi$. In each ResBlock, the time embedding and the parameters embedding are then transformed by a small MLP with a depth of 1, prior to being added to the input of the respective ResBlock.

In addition, for the cosmological component separation task only, we trained a model exclusively on low noise levels by reducing the value of $\beta_\mathrm{max}$ to $2$, as opposed to $20$. Consequently, our network is limited to denoising images for $\sigma$ values within the specified prior range. This adjustment restricts the model's utility as a generative prior but enhances its precision in reconstructing summary statistics and calibrating the inference method. Furthermore, we found that at lower noise-levels, feeding the network Fourier features~\citep[as in][]{kingma2021on} improved calibration of the posterior distribution.

\subsection{Loss Functions}
Our models are trained by minimizing a modified denoising score-matching loss whose original formulation would be:
\begin{equation}
\label{eqapp:score_matching_loss}
     \mathcal{L}(\ve{\theta}) = \mathbb{E}_{t,\phi}\left[~\lambda(t)\mathbb{E}_{\ve{z}_0,\,\ve{z}_t\vert(\ve{z}_0,\ve{\phi})}\left[\left\lVert \mathbf{s}_\ve{\theta}(\ve{z}_t,t)+\mathbf{\Sigma_\ve{\phi}}^{-1}\left(\ve{z}_t-\sqrt{\bar\alpha(t)}\ve{z}_0\right)/\left(1-\bar\alpha(t)\right)\right\lVert^2_2\right]\right],
\end{equation}
with $\bar\alpha(t)=\exp(-\int_0^t\beta(u)\mathrm{d}u)$. As in \citet{Heurtel2023}, we re-normalize all vectors in this loss by multiplying them by $- \sqrt{1-\bar\alpha(t)}\mathbf{\Sigma_\phi}$ to stabilize the learning process and reduce numerical errors. Consequently, the adjusted loss function read:
\begin{equation}\label{eq:reweighted_score_matching_loss}
    \mathcal{L}(\ve{\theta})=\mathbb{E}_{t,\phi}\left\{~\lambda(t)\mathbb{E}_{\ve{z}_0,\,\ve{z}_t\vert(\ve{z}_0,\phi)}\left[\left\lVert \mathbf{m}_\ve{\theta}(\ve{z}_t,t,\phi)-\frac{1}{\sqrt{1-\bar\alpha(t)}} \left(\ve{z}_t-\sqrt{\bar\alpha(t)}\ve{z}_0\right)\right\lVert^2_2\right]\right\}
\end{equation}
This loss in turns corresponds to the standard loss from \citet{ho2020denoising} where one tries to estimate the added noise (up to a rescaling constant).

In order to verify that the model is not overfitting, we check for mode collapse and dataset copying with a specific $L_2$ distance taking into account the periodic boundary conditions. This was particularly important as our models seem more prone to overfitting than model trained under white noise.

\section{Properties of the Gibbs sampler} \label{app:Gibbs}

In this section, we derive theoretical properties of the proposed Gibbs sampler.
Specifically, we work out the sampler's stationary distribution, review conditions for its existence, and quantify its error.

In what follows, all distributions are conditioned on $\ve{y}$, and so we drop this dependency to alleviate the notation.

\subsection{Formal notation}

Any Markov chain Monte Carlo (MCMC) sampler defines a transition kernel between a current state $(\phi, \ve{x})$ and a new state $(\phi', \ve{x}')$.
We denote this transition $T((\phi, \ve{x}) \to (\phi', \ve{x}'))$.
The \textit{invariant distribution} $\pi$ is the distribution that verifies
\begin{equation} \label{eq:invariant}
    \int T((\phi, \ve{x}) \to (\phi', \ve{x}')) \pi(\phi, \ve{x}) \text 
\ d \phi \text d \ve{x} = \pi(\phi', \ve{x}').
\end{equation}
In words, if we apply the transition kernel $T$ to a state $(\phi, \ve{x})$ drawn from $\pi$, then the new state $(\phi', \ve{x}')$ is also distributed according to $\pi$.

The transition kernel $T$ of the Gibbs sampler decomposes into two kernels $T = T_\phi \circ T_\ve{x}$.
The first transition kernel $T_\phi$ updates $\phi$ but maintains $\ve{x}$ fixed, while $T_\ve{x}$ updates $\ve{x}$ and leaves $\phi$ unchanged.

$T_\phi$ draws $\phi$ from an MCMC sampler that admits $p(\phi \mid \ve{x}, \ve{y})$ as its stationary distribution.
In our paper, we use Hamiltonian Monte Carlo but the results we derive here would hold for any MCMC algorithm with the right stationary distribution.
The MCMC sampler defines a transition density $\tau(\phi \to \phi' \mid \ve{x})$, and moreover
\begin{equation} \label{eq:Tphi}
    T_\phi = \tau(\phi \to \phi' \mid \ve{x}) \delta(\ve{x}'=\ve{x}),
\end{equation}
where $\delta$ is the dirac delta function.
$\tau$ may correspond to the transition obtained after applying either a single or multiple iterations of the MCMC sampler.

Next, $T_\ve{x}$ draws samples from a diffusion model that approximates the data generating process $p(\ve{x} \mid \phi, \ve{y})$.
It what follows, we assume that the distribution of the samples produced by the diffusion model admits a probability density $q(\ve{x} \mid \phi, \ve{y})$.
Then
\begin{equation} \label{eq:Tx}
  T_\ve{x} = q(\ve{x}' \mid \phi) \delta(\phi'=\phi).
\end{equation}

\subsection{Proof of Proposition~\ref{prop:invariant}: invariant distributions}

\begin{proof}
  Let $\pi(\phi, \ve{x})$ be any joint distribution, such that $\pi(\phi \mid \ve{x}) = p(\phi \mid \ve{x})$.
  Then
  \begin{eqnarray}
    \int T_\phi ((\phi, \ve{x}) \to (\phi', \ve{x}')) \pi(\phi, \ve{x}) \text d \phi \text d \ve{x} 
        & = & \int \tau(\phi \to \phi' \mid \ve{x}) \delta(\ve{x}' = \ve{x}) \pi(\phi, \ve{x}) \text d \phi \text d \ve{x} \nonumber \\
        & = & \int \tau(\phi \to \phi' \mid \ve{x}') p(\phi \mid \ve{x}') \pi(\ve{x}') \text d \phi \nonumber \\
        & = & \pi(\ve{x}') \int \tau(\phi \to \phi' \mid \ve{x}') p(\phi \mid \ve{x}') \text d \phi \nonumber \\
        & = & \pi(\ve{x}') p(\phi \mid \ve{x}') \nonumber \\
        & = & \pi(\phi', \ve{x}'),
  \end{eqnarray}
  where for the penultimate line we used the assumption that $\tau$'s stationary distribution is $p(\phi \mid \ve{x})$.
  From \ref{eq:invariant}, we then have that $\pi(\phi, \ve{x})$ is an invariant distribution of $T_\phi$.

  Similarly let $\tilde \pi(\phi, \ve{x})$ be any joint distribution such that $\tilde \pi(\ve{x} \mid \phi) = q(\ve{x} \mid \phi)$.
  Then
  \begin{eqnarray}
      \int T_\ve{x}((\phi, \ve{x}) \to (\phi', \ve{x}')) \tilde \pi(\phi, \ve{x}) \text d \phi \text d \ve{x} 
        & = & \int q(\ve{x}' \mid \phi) \delta(\phi' = \phi) \tilde \pi(\phi, \ve{x}) \text d \phi \text d \ve{x} \nonumber \\
        & = & q(\ve{x}' \mid \phi') \int  \tilde \pi(\phi', \ve{x}) \text d \ve{x} \nonumber \\
        & = & q(\ve{x}' \mid \phi') \tilde \pi(\phi') \nonumber \\
        & = & \tilde \pi(\phi', \ve{x}').
  \end{eqnarray}
  Hence $\tilde \pi(\phi, \ve{x})$ is invariant under $T_\ve{x}$.
\end{proof}

\subsection{Existence of Stationary Distribution}

\begin{definition}
  Consider two families of conditional distributions, $f(\phi \mid \ve{x})$ and $g(\phi \mid \ve{x})$.
  We say $f$ and $g$ are \textit{compatible} if there exists a joint distribution $\pi(\phi, \ve{x})$ such that $\pi(\phi \mid \ve{x}) = f(\phi \mid \ve{x})$ and $\pi(\ve{x} \mid \phi) = g(\ve{x} \mid \phi)$.
  Else, we say that $f$ and $g$ are \textit{incompatible}.
\end{definition}
More general definitions of compatibility can be found in references \citep{Arnold2001, Liu2012}.
Sufficient conditions for compatibility are provided by \citet[Theorem 1]{Arnold2001}, restated here for convenience (and adapted to our notation).\footnote{As argued by \citet[Theorem~2.3]{Liu:2021}, the conditions by \citet{Arnold2001} are not necessary and item 1 in \Cref{thm:compatible} can be replaced with a weaker condition.}
\begin{theorem} \label{thm:compatible}
  $f(\phi \mid \ve{x})$ and $g(\ve{x} \mid \phi)$ are compatible if
  \begin{enumerate}
      \item $\{(\phi, \ve{x}): f(\phi \mid \ve{x}) > 0\} = \{(\phi, \ve{x}): g(\ve{x} \mid \phi) > 0\} := S$,

      \item There exist functions $u(\ve{x})$ and $v(\phi)$ such that for every $(\phi, \ve{x}) \in S$,
      \begin{equation}
          \frac{g(\ve{x} \mid \phi)}{f(\phi \mid \ve{x})} = u(\ve{x}) v(\phi),
      \end{equation}
      with $u(\ve{x})$ integrable, that is
      \begin{equation} \label{eq:integrable}
          \int_{\mathcal{X}} u(\ve{x}) \text d \ve{x} < \infty.
      \end{equation}
  \end{enumerate}
\end{theorem}
We find it enlightening to rewrite the last condition as an integral over the left-hand-side of Eq.~\eqref{eq:integrable}.
Applying this condition to our problem we have
\begin{equation*}
    \int_\mathcal{X} \frac{q(\ve{x} \mid \phi, \ve{y})}{p(\phi \mid \ve{x}, \ve{y})} \text d \ve{x} < \infty,
\end{equation*}
which is Eq.~\eqref{eq:compatible}.

\subsection{Error in the Stationary Distribution}

Moving forward, we assume that $p(\phi \mid \ve{x})$ and $q(\ve{x} \mid \phi)$ are compatible, and so our Gibbs sampler admits a stationary distribution $\pi(\ve{x}, \phi)$.
We now prove Theorem~\ref{thm:KL}, which tells us the KL-divergence from the true (posterior) distribution $p(\phi)$ to the marginal stationary distribution $\pi(\phi)$ of our Gibbs sampler.

\begin{proof}
  Standard application of the probability chain rule gives
  \begin{equation}
      p(\ve{x}) = p(\phi) \frac{p(\ve{x} \mid \phi)}{p(\phi \mid \ve{x})}.
  \end{equation}
  Clearly, $p(\phi)$ must act as a normalizing constant and so
  \begin{equation}
    p(\phi) = \frac{1}{\int_\mathcal{X} \frac{p(\ve{x} \mid \phi)}{p(\phi \mid \ve{x})} \text d \ve{x}}.
  \end{equation}
  Recalling that $\pi(\phi \mid \ve{x}) = p(\phi \mid \ve{x})$, a similar argument yields
  \begin{equation}
      \pi(\phi) = \frac{1}{\int_\mathcal{X} \frac{q(\ve{x} \mid \phi)}{p(\phi \mid \ve{x})} \text d \ve{x}}.
  \end{equation}
  Then
  \begin{eqnarray} \label{eq:density_ratio}
    & \frac{1}{p(\phi)} - \frac{1}{\pi(\phi)} & = \int_\mathcal{X} \frac{p(\ve{x} \mid \phi) - q(\ve{x} \mid \phi)}{p(\phi \mid \ve{x})} \text d \ve{x} \nonumber \\
    \iff & \frac{\pi(\phi) - p(\phi)}{p(\phi) \pi(\phi)} & = \int_\mathcal{X} \frac{p(\ve{x} \mid \phi) - q(\ve{x} \mid \phi)}{p(\phi \mid \ve{x})} \text d \ve{x} \nonumber \\
    \iff & \frac{\pi(\phi) - p(\phi)}{\pi(\phi)} & = \int_\mathcal{X} (p(\ve{x} \mid \phi) - q(\ve{x} \mid \phi)) \frac{p(\ve{x})}{p(\ve{x} \mid \phi)} \text d \ve{x} \nonumber \\
    \iff & 1 - \frac{p(\phi)}{\pi(\phi)} & = 1 - \int_\mathcal{X} \frac{q(\ve{x} \mid \phi)}{p(\ve{x} \mid \phi)} p(\ve{x}) \text d \ve{x} \nonumber \\
    \iff & \frac{p(\phi)}{\pi(\phi)} & = \int_\mathcal{X} \frac{q(\ve{x} \mid \phi)}{p(\ve{x} \mid \phi)} p(\ve{x}) \text d \ve{x}.
  \end{eqnarray}
  %
  Finally recall that
  \begin{equation}
      \text{KL}(p(\phi) || \pi(\phi)) = \int_\Phi \log \frac{p(\phi)}{\pi(\phi)} p(\phi) \text d \phi,
  \end{equation}
  and so taking on both sides of Eq.~\eqref{eq:density_ratio} the logarithm and the expectation value with respect to $p(\phi)$, we obtain the wanted result.
\end{proof}

\textit{Remark.} Eq.~\eqref{eq:density_ratio}, which applies to the density ratio at any point, provides a more general result than the result on averaged density ratios stated in Theorem~\ref{thm:KL}.

\section{Limitations of Guidance-based Posterior Sampling Methods}
\label{app:other_methods}

By examining the applicability of existing methods DPS \cite{chung2023diffusion} and $\Pi$-GDM \citep{Song2023} in the cosmology application presented in Sect.~\ref{sec:cosmo_images}, we identified limitations in these guidance-based methods, which further legitimize the interest in our approach. We report our investigation in this appendix.

First, it is important to note that, since the exact conditional guidance term is typically intractable, guidance-based posterior sampling methods require methodological approximations. These approximations may impact the estimation of the posterior distribution in unpredictable ways. For scientific applications like our problem of cosmological inference from CMB data, this can be a significant downside. Furthermore, we encountered significant numerical difficulties with DPS and $\Pi$-GDM, which we consider strong evidence of the limitations of these methods for our cosmology application.

Let us first rewrite Eq.~\eqref{eq:denoising_pb} as a standard linear inverse problem equation (i.e., involving white Gaussian noise). Starting from Eq.~\eqref{eq:denoising_pb}:
\begin{equation}
    \ve{y} = \ve{x} + \ve{\varepsilon},\quad\text{with} \; \ve{\varepsilon}\sim\mathcal{N}(\ve{0},\ve{\Sigma}_\ve{\phi}),
\end{equation}
and assuming $\ve{\Sigma}_\ve{\phi}$ is invertible, we define $\tilde{\ve{\varepsilon}} = \ve{\Sigma}_\ve{\phi}^{-1/2}\ve{\varepsilon}$, so that $\tilde{\ve{\varepsilon}}\sim\mathcal{N}(\ve{0},\ve{I}_d)$. Then, introducing $\tilde{\ve{y}}_{\ve{\phi}} = \ve{\Sigma}_\ve{\phi}^{-1/2}\ve{y}$ and $\ve{A}_{\ve{\phi}} = \ve{\Sigma}_\ve{\phi}^{-1/2}$, we get:
\begin{equation}
    \tilde{\ve{y}}_{\ve{\phi}} = \ve{A}_{\ve{\phi}} \ve{x} + \tilde{\ve{\varepsilon}},
\end{equation}
which resembles a standard linear inverse problem where the linear operator is $\ve{A}_{\ve{\phi}} = \ve\Sigma_\ve{\phi}^{-1/2}$. Also, note that the observation $\tilde{\ve{y}}_{\ve{\phi}}$ depends on the parameters $\ve{\phi}$, which adds a layer of complexity when the goal is to infer $\ve{\phi}$.

\begin{figure*}
    \includegraphics[width=0.95\textwidth]{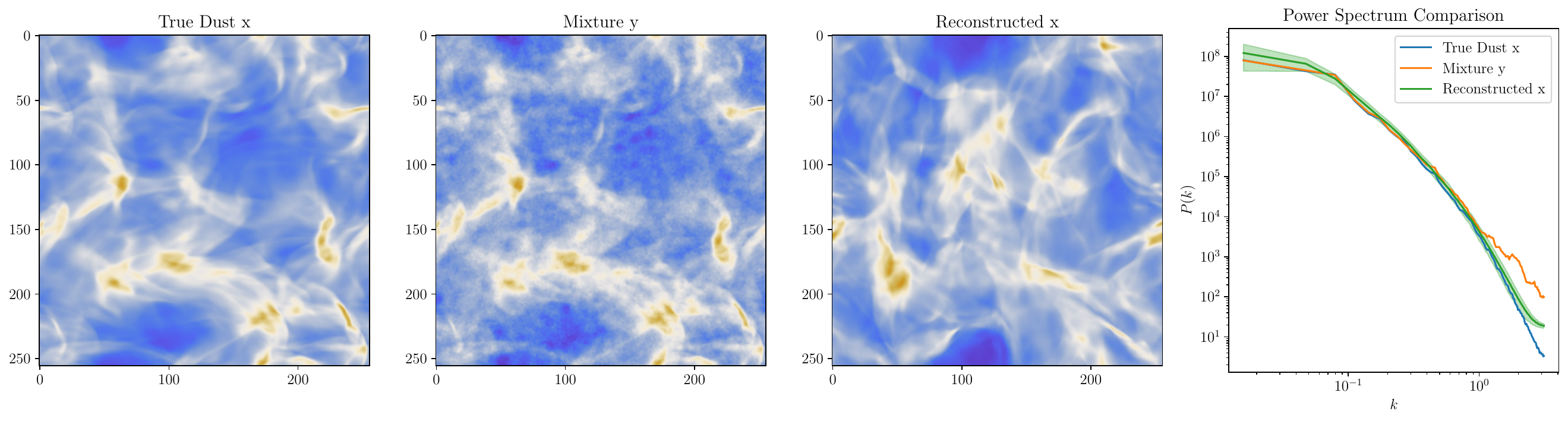}
    \caption{Best results when optimizing the guidance strength for the standard off-the shelf $\Pi$-GDM algorithm.}
    \label{fig:pigdm_best_guidance}
\end{figure*}

\begin{figure*}
    \includegraphics[width=0.95\textwidth]{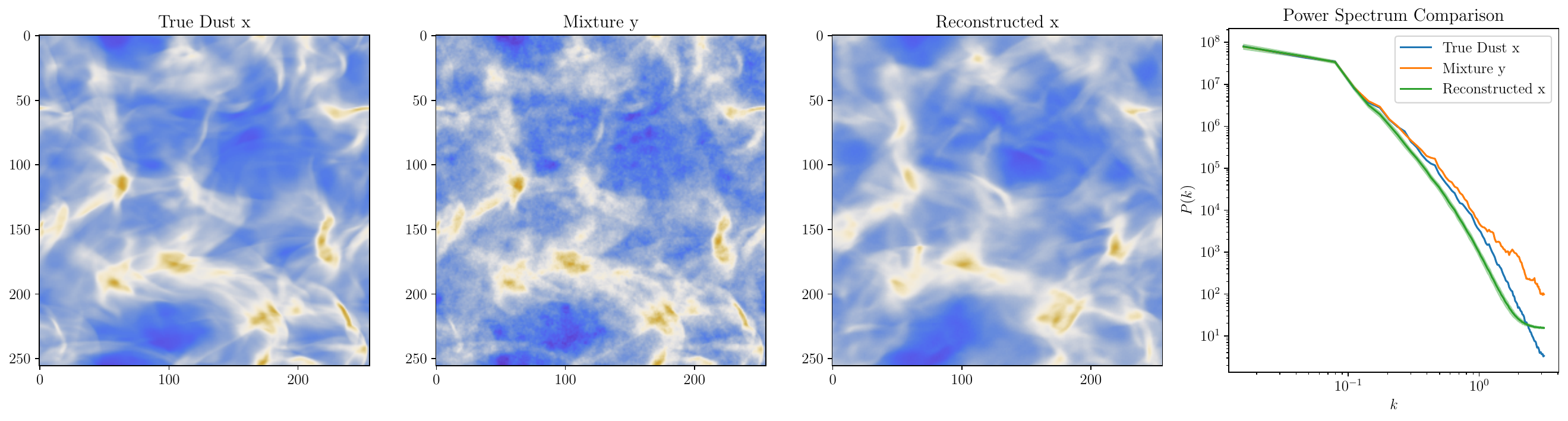}
    \caption{Best results when optimizing a regularization in a modified $\Pi$-GDM.}
    \label{fig:pigdm_best_features}
\end{figure*}

We implemented off-the-shelf DPS \cite{chung2023diffusion} and $\Pi$-GDM \citep{Song2023} algorithms in the context of our cosmology application\footnote{Note that, in our case, the $\Pi$-GDM guidance term is straightforward to compute since the operator $\ve\Sigma_\ve{\phi}^{-1/2}$ is diagonal in Fourier space.}. The standard regularization (guidance strength) schemes initially yielded numerical float overflows. To avoid these numerical limitations, we had to decrease the guidance strength to a point where it was clear that we were not sampling the posterior distribution (large scales structure were off, as shown in Fig.~\ref{fig:pigdm_best_guidance}). Our most satisfying approach turned out to be a variation of $\Pi$-GDM involving a time-dependent regularization term~\citep[][App. A.3]{Song2023} way outside the normal range for any inverse problems. Fig.~\ref{fig:pigdm_best_features} depicts our results with associated summary statistics, proving that the posterior distribution is very poorly estimated. Furthermore, when trying to use this last method in our Gibbs pipeline, we could not reach convergence to any posterior distribution over $\phi$.

We interpret these numerical challenges as the result of the significant dynamic range of eigenvalues of $\ve\Sigma_\ve{\phi}$. For our cosmology application, the condition number of $\ve\Sigma_\ve{\phi}$ is approximately $6\times10^5$, and the largest eigenvalue of $\ve\Sigma_\ve{\phi}^{-1/2}$ is $1.6\times10^2$. In comparison, standard inverse problems such as inpainting, deblurring, or super-resolution problems, typically involve operators with eigenvalues bounded between 0 and 1. Since guidance-based methods rely on estimates of $\nabla_{\ve{x}_t}\log p(\ve x_t\vert \ve{y}, \ve{\phi})$, these large eigenvalues lead to numerical instabilities along their corresponding eigenvectors (i.e., for our problem, at high frequencies). If we attempt to mitigate this instability using regularization, the guidance along eigenvectors linked to the smallest eigenvalues of $\ve{\Sigma}_\ve{\phi}^{-1/2}$ becomes insufficient, resulting in impaired low frequencies/large-scale structures.

\section{Additional Results}

\begin{figure*}
    \centering
    \raisebox{2em}{\includegraphics[width=0.72\hsize]{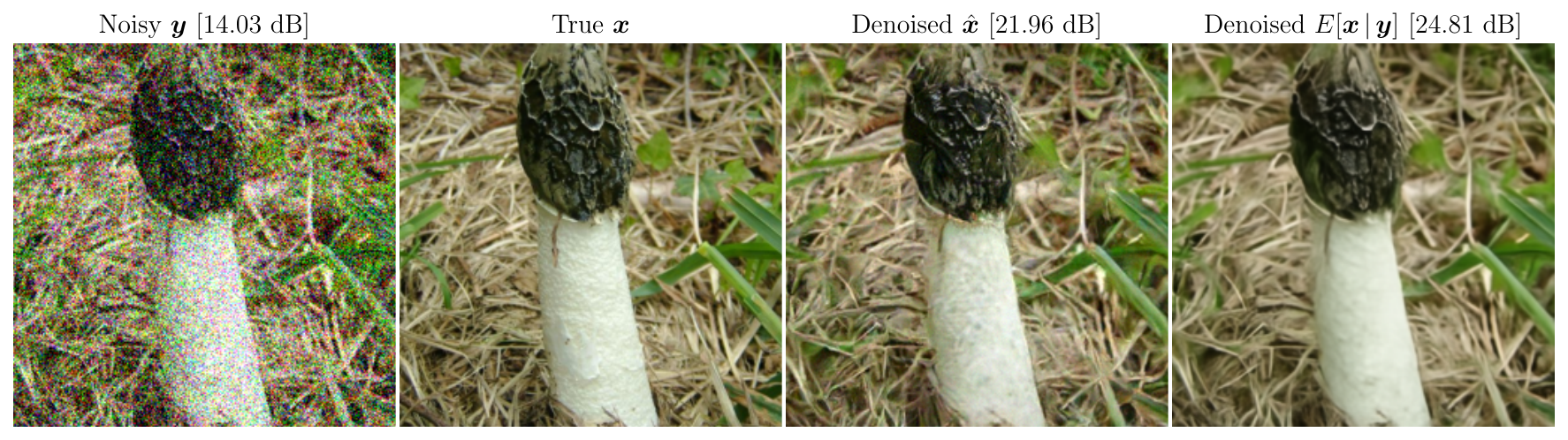}}
    \includegraphics[width=0.25\hsize]{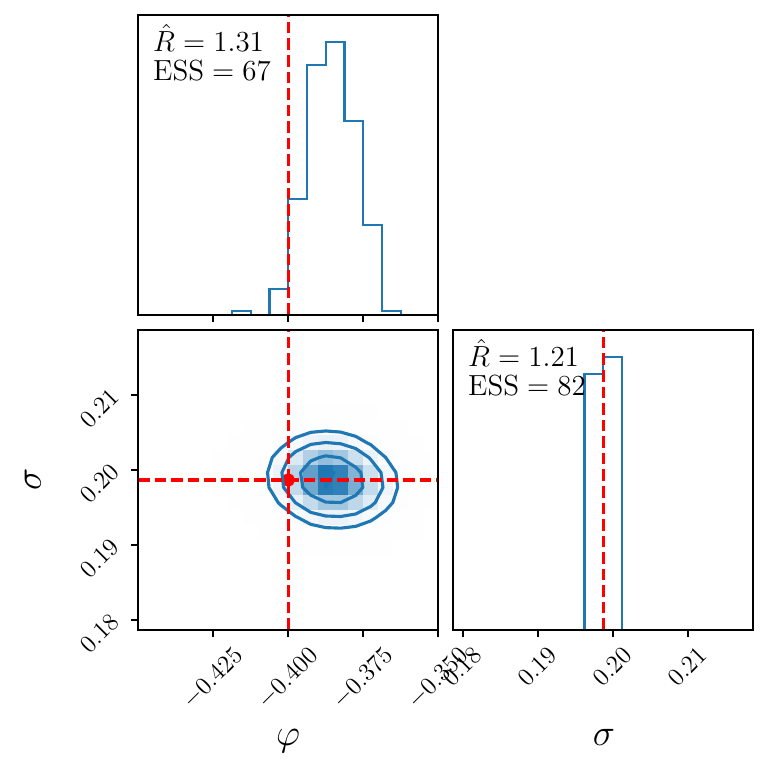}
    \raisebox{2em}{\includegraphics[width=0.72\hsize]{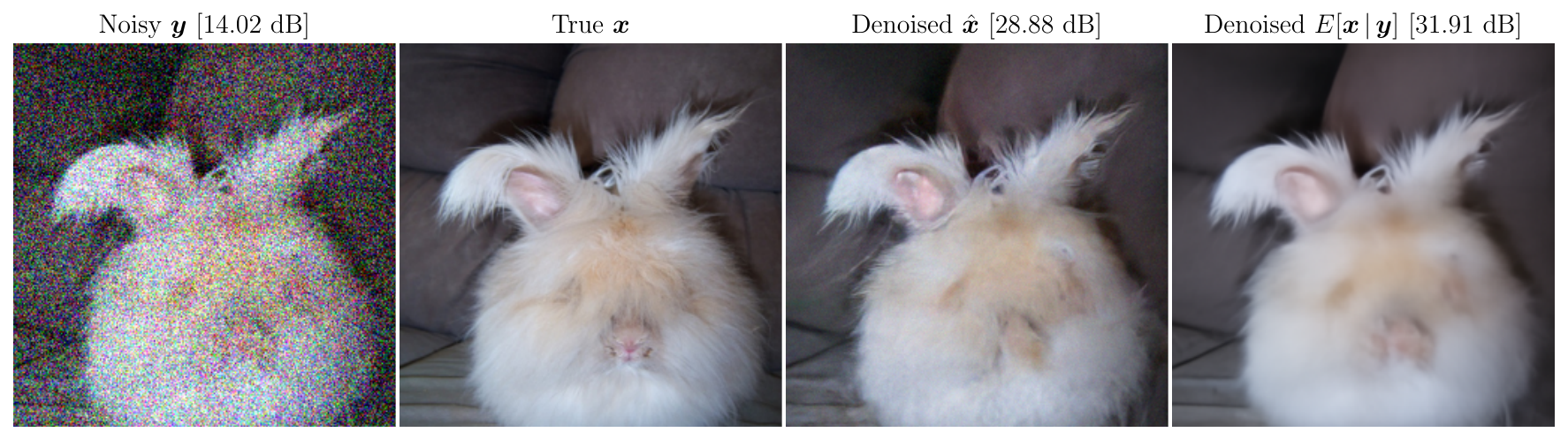}}
    \includegraphics[width=0.25\hsize]{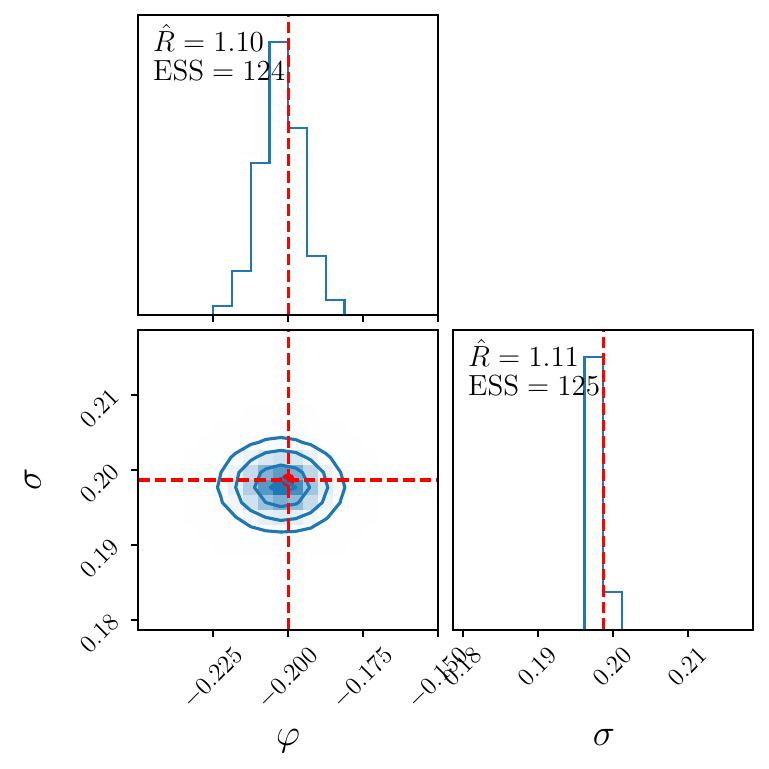}
    \raisebox{2em}{\includegraphics[width=0.72\hsize]{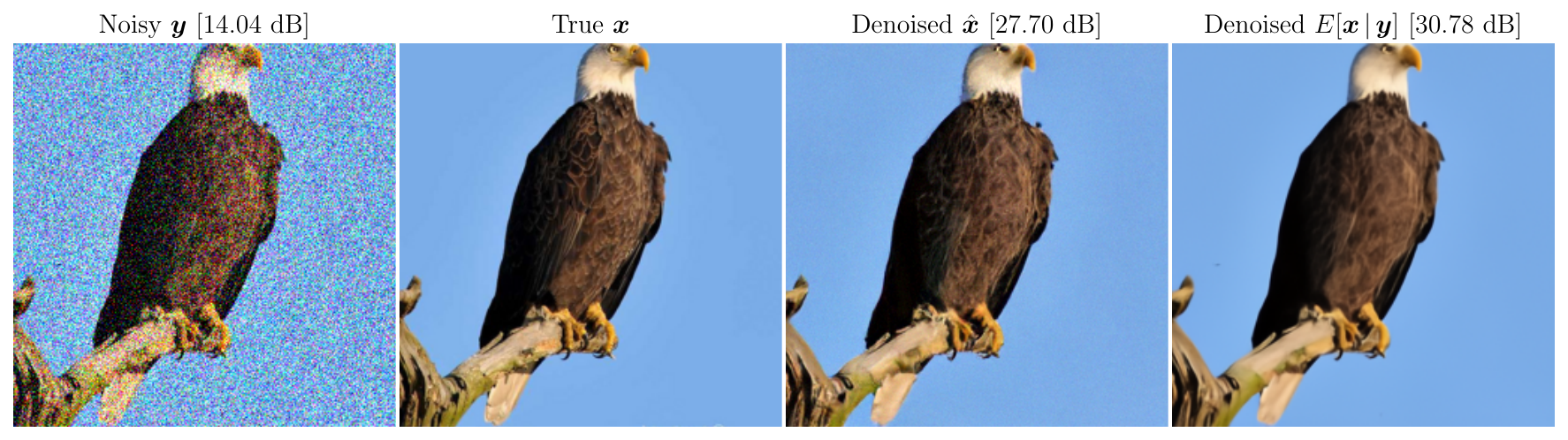}}
    \includegraphics[width=0.25\hsize]{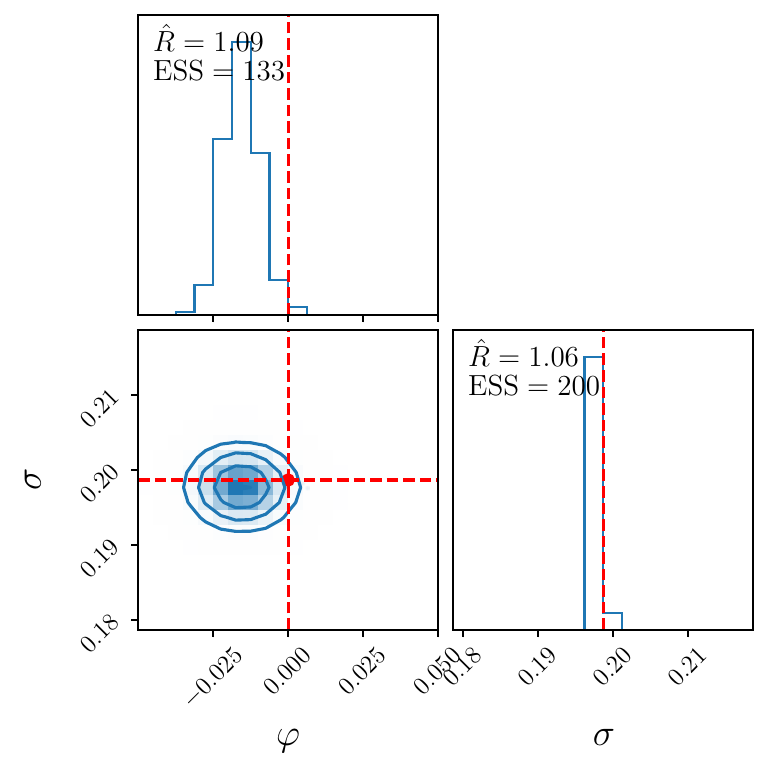}
    \raisebox{2em}{\includegraphics[width=0.72\hsize]{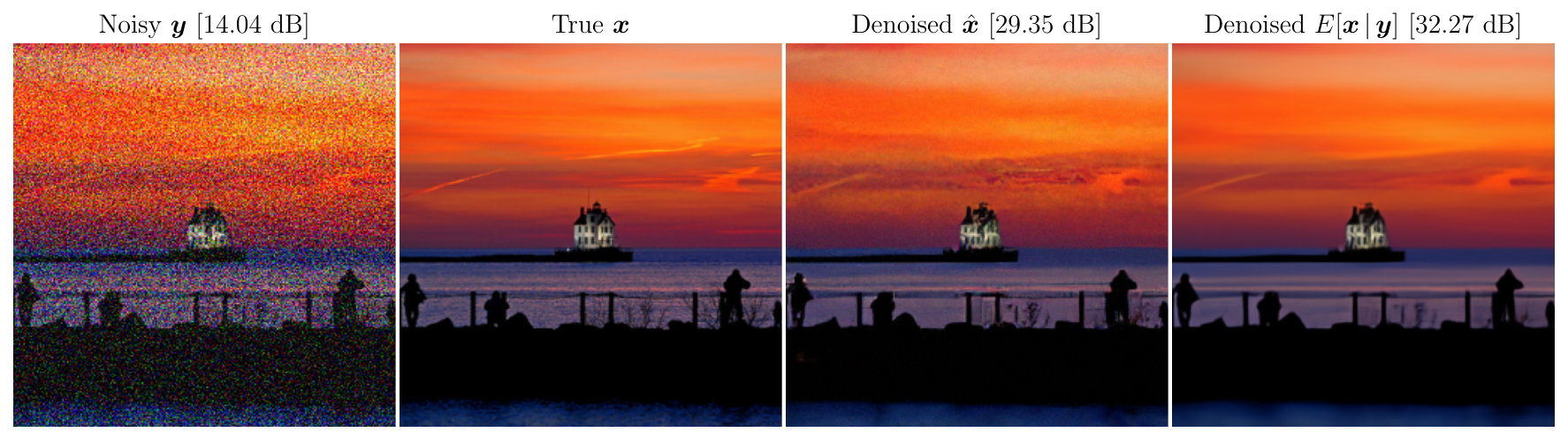}}
    \includegraphics[width=0.25\hsize]{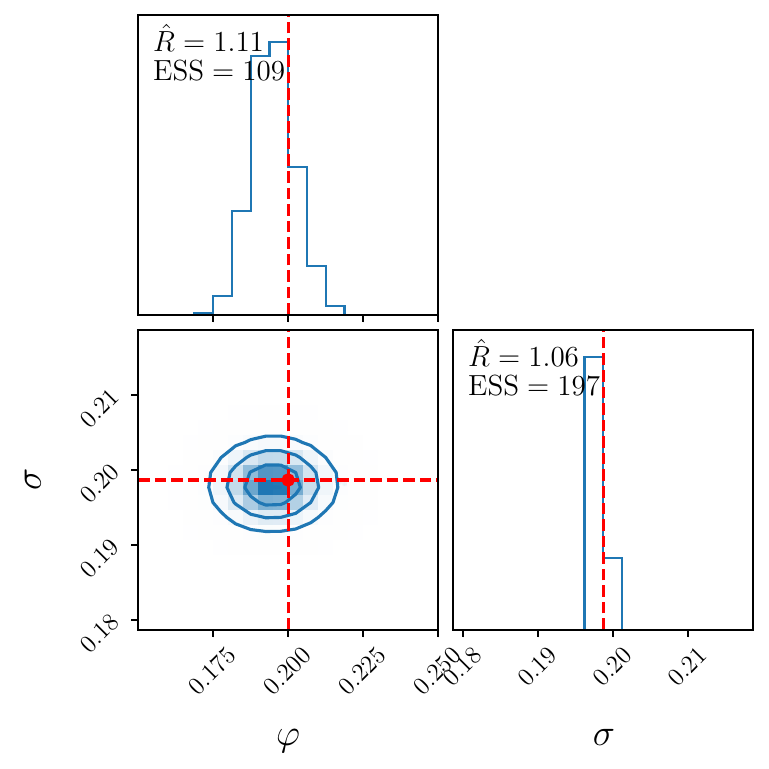}
    \raisebox{2em}{\includegraphics[width=0.72\hsize]{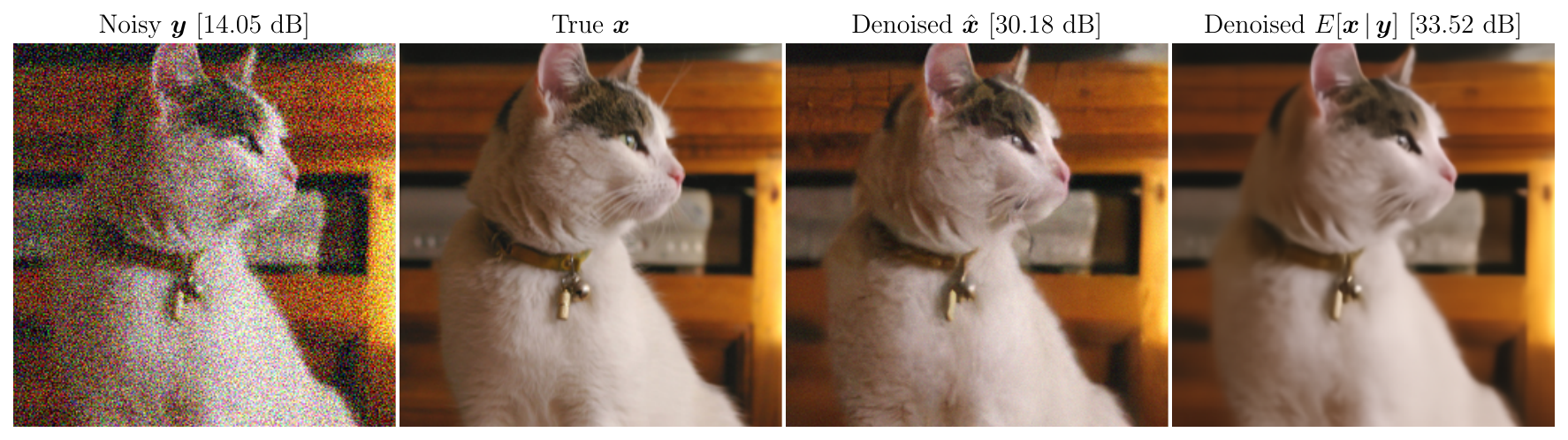}}
    \includegraphics[width=0.25\hsize]{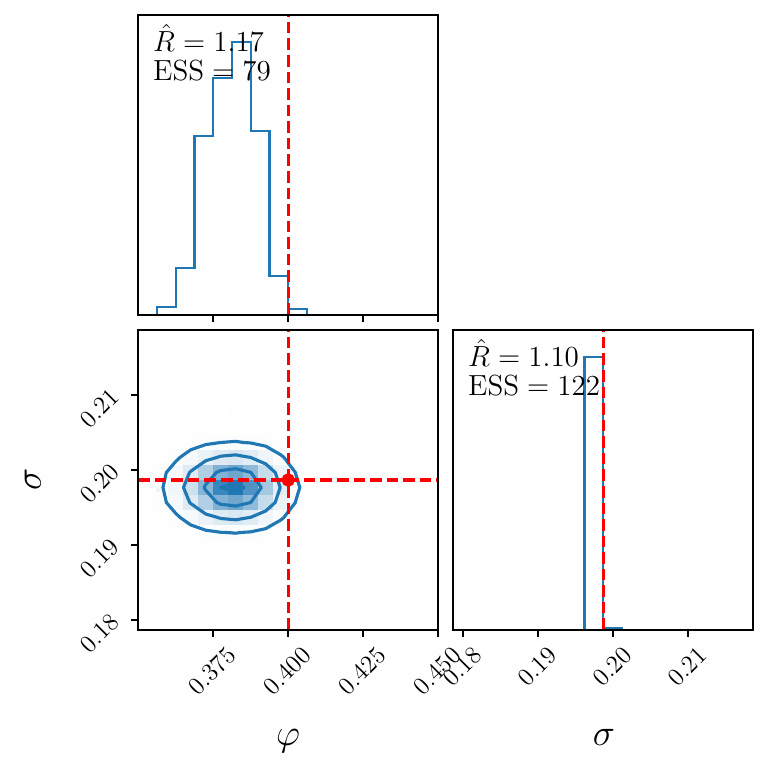}
    \caption{Examples of blind denoising with GDiff on noisy ImageNet mocks with $\sigma = 0.2$ and $\varphi \in \{-0.4, -0.2, 0.0, 0.2, 0.4\}$ (from top to bottom). \textit{Left:} Noisy examples $\ve{y}$ next to the noise-free images $\ve{x}$, denoised samples $\hat{\ve{x}}$ and estimates of the posterior mean $\Eb{\ve{x}\,|\,\ve{y}}$. \textit{Right:} Inferred posterior distributions over the noise parameters.}
    \label{fig:nat_images_denoising_other_examples}
\end{figure*}

\begin{table}[H]
{\tiny 
\setlength{\tabcolsep}{2pt} 
\centering
\begin{tabular}{ c c || c c c c || c c c c || c c c c}
\multicolumn{1}{c}{\multirow{3}{*}{Dataset}} & \multicolumn{1}{c||}{\multirow{3}{*}{\parbox{1cm}{\centering Noise Level $\sigma$}}} &  \multicolumn{4}{c||}{$\varphi = -1 \rightarrow$ {\color{betterpink} Pink} noise} & \multicolumn{4}{c||}{$\varphi = 0 \rightarrow $ White noise} & \multicolumn{4}{c}{$\varphi = 1 
 \rightarrow$ {\color{blue} Blue} noise}\\
\cline{3-14}
& & \multirow{2}{*}{BM3D} & \multirow{2}{*}{DnCNN} & \multirow{2}{*}{\parbox{1cm}{\centering \textbf{GDiff} $~~\hat{\ve{x}}~~$}} & \multirow{2}{*}{\parbox{1.3cm}{\centering \textbf{GDiff} $\Eb{\ve{x}\,|\,\ve{y}}$ }} & \multirow{2}{*}{BM3D} & \multirow{2}{*}{DnCNN} & \multirow{2}{*}{\parbox{1cm}{\centering \textbf{GDiff} $~~\hat{\ve{x}}~~$}} & \multirow{2}{*}{\parbox{1.3cm}{\centering \textbf{GDiff} $\Eb{\ve{x}\,|\,\ve{y}}$ }} & \multirow{2}{*}{BM3D} & \multirow{2}{*}{DnCNN} & \multirow{2}{*}{\parbox{1cm}{\centering \textbf{GDiff} $~~\hat{\ve{x}}~~$}} & \multirow{2}{*}{\parbox{1.3cm}{\centering \textbf{GDiff} $\Eb{\ve{x}\,|\,\ve{y}}$ }}\\
&&&&&&&&&&&&& \\
\hline
\multirow{3}{*}{ImageNet} & 0.06 & \mes{0.90}{0.01} & \mes{(0.88)}{0.00} & \mes{0.86}{0.01} & \textbf{\mes{0.92}{0.00}} & \mes{0.92}{0.00} & \mes{0.92}{0.00} & \mes{0.88}{0.01} & \textbf{\mes{0.93}{0.00}} & \mes{0.94}{0.00} & \mes{(0.92)}{0.00} & \mes{0.90}{0.00} & \textbf{\mes{0.95}{0.00}} \\ 
 & 0.1 & \mes{0.81}{0.01} & \mes{(0.76)}{0.01} & \mes{0.77}{0.01} & \textbf{\mes{0.86}{0.01}} & \mes{0.90}{0.01} & \mes{0.90}{0.01} & \mes{0.84}{0.01} & \textbf{\mes{0.91}{0.00}} & \mes{0.90}{0.01} & \mes{(0.88)}{0.01} & \mes{0.85}{0.01} & \textbf{\mes{0.92}{0.00}} \\ 
 & 0.2 & \mes{0.63}{0.01} & \mes{(0.53)}{0.01} & \mes{0.62}{0.02} & \textbf{\mes{0.74}{0.02}} & \mes{0.79}{0.01} & \mes{0.80}{0.01} & \mes{0.74}{0.01} & \textbf{\mes{0.83}{0.01}} & \mes{0.83}{0.01} & \mes{(0.79)}{0.01} & \mes{0.78}{0.01} & \textbf{\mes{0.87}{0.01}} \\
 \hline
 \multirow{3}{*}{CBSD68} & 0.06 & \mes{0.90}{0.01} & \mes{(0.88)}{0.00} & \mes{0.84}{0.01} & \textbf{\mes{0.92}{0.00}} & \mes{0.93}{0.00} & \textbf{\mes{0.94}{0.00}} & \mes{0.88}{0.00} & \textbf{\mes{0.94}{0.00}} & \mes{0.94}{0.00} & \mes{(0.94)}{0.00} & \mes{0.90}{0.00} & \textbf{\mes{0.95}{0.00}} \\ 
 & 0.1 & \mes{0.82}{0.01} & \mes{(0.78)}{0.01} & \mes{0.75}{0.01} & \textbf{\mes{0.85}{0.01}} & \mes{0.89}{0.00} & \textbf{\mes{0.90}{0.00}} & \mes{0.81}{0.01} & \textbf{\mes{0.90}{0.00}} & \mes{0.91}{0.00} & \mes{(0.90)}{0.00} & \mes{0.86}{0.00} & \textbf{\mes{0.93}{0.00}} \\ 
 & 0.2 & \mes{0.65}{0.01} & \mes{(0.55)}{0.01} & \mes{0.61}{0.01} & \textbf{\mes{0.74}{0.01}} & \mes{0.79}{0.01} & \mes{0.80}{0.01} & \mes{0.69}{0.01} & \textbf{\mes{0.81}{0.01}} & \mes{0.83}{0.01} & \mes{(0.79)}{0.01} & \mes{0.76}{0.01} & \textbf{\mes{0.86}{0.01}} \\ 

\end{tabular}
}
\caption{Image quality in terms of SSIM ($\uparrow$) after denoising with GDiff (blind) and baselines BM3D (non-blind) and DnCNN (blind). We report mean PSNR and standard error computed on batches of 50 images. For GDiff, we provide performance for both posterior samples $\ve{x}$ and estimates of the posterior mean $\Eb{\ve{x}\,|\,\ve{y}}$. We point out that DnCNN was trained with white noises only, hence results obtained for $\varphi \neq 0$ could be sub-optimal.}
\label{table:natimages_denoising_ssim_with_errorbar}
        \vspace{-1em}
\end{table}

\begin{figure*}
    \includegraphics[width=0.49\hsize]{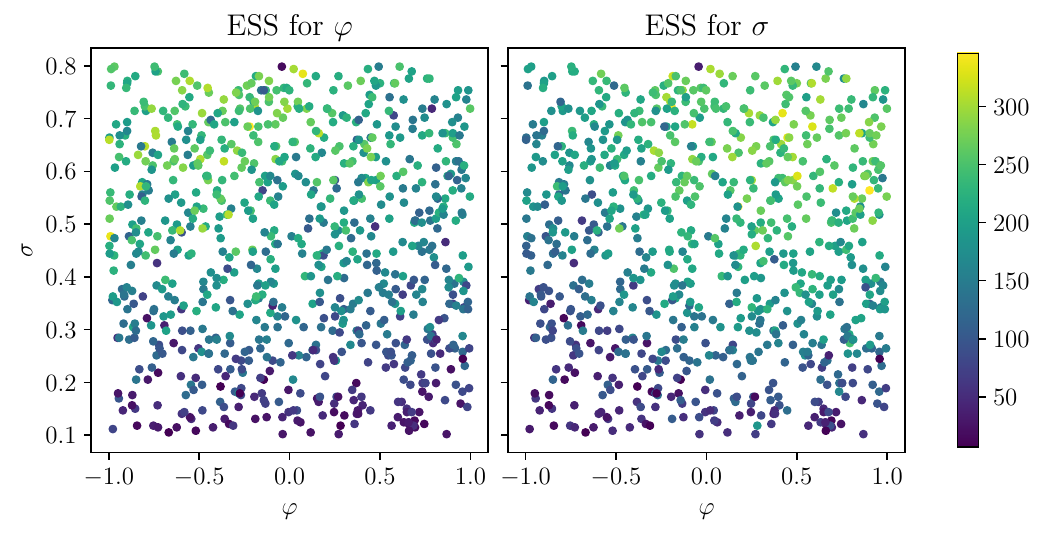}
    \includegraphics[width=0.49\hsize]{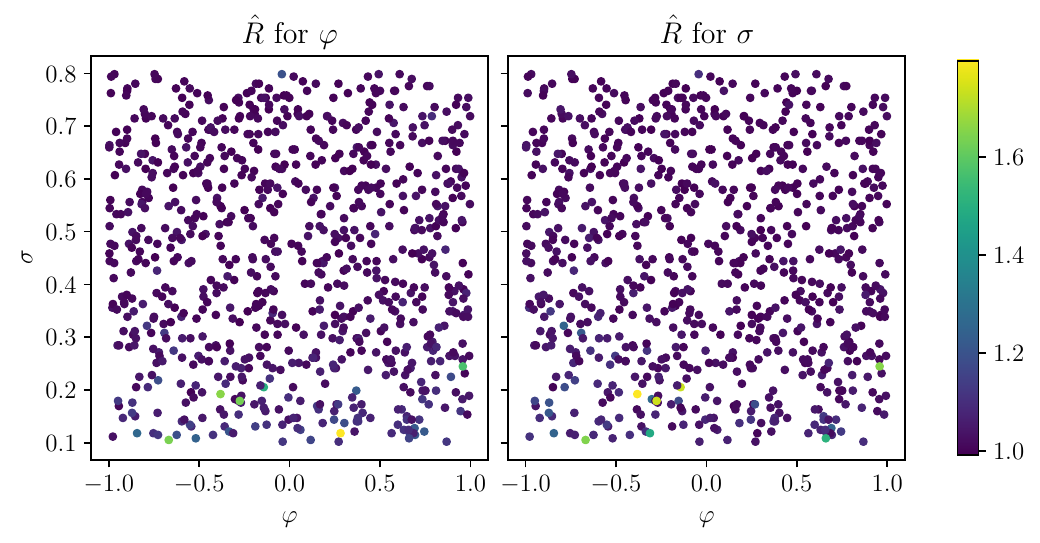}
    \caption{ESS (left) and $\hat{R}$ (right) statistics for inferences on noisy natural images across the parameter space.}
    \label{fig:nat_images_validation}
\end{figure*}

\begin{figure*}
    \centering
    \includegraphics[width=0.8\hsize]{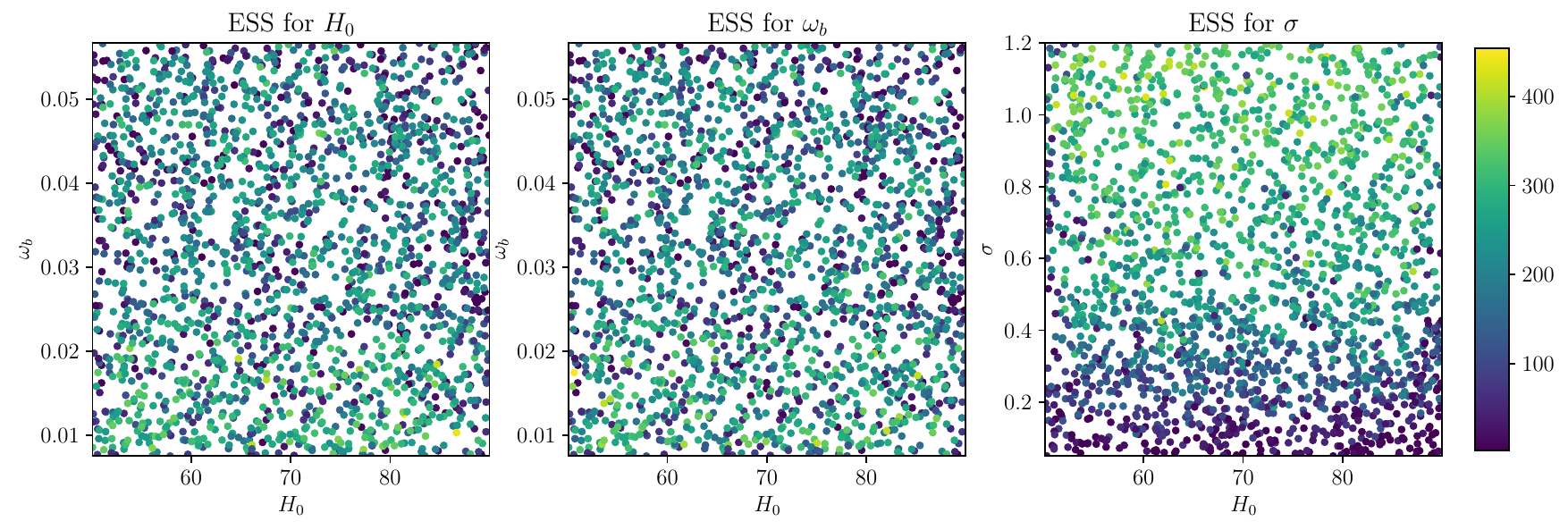}
    \includegraphics[width=0.8\hsize]{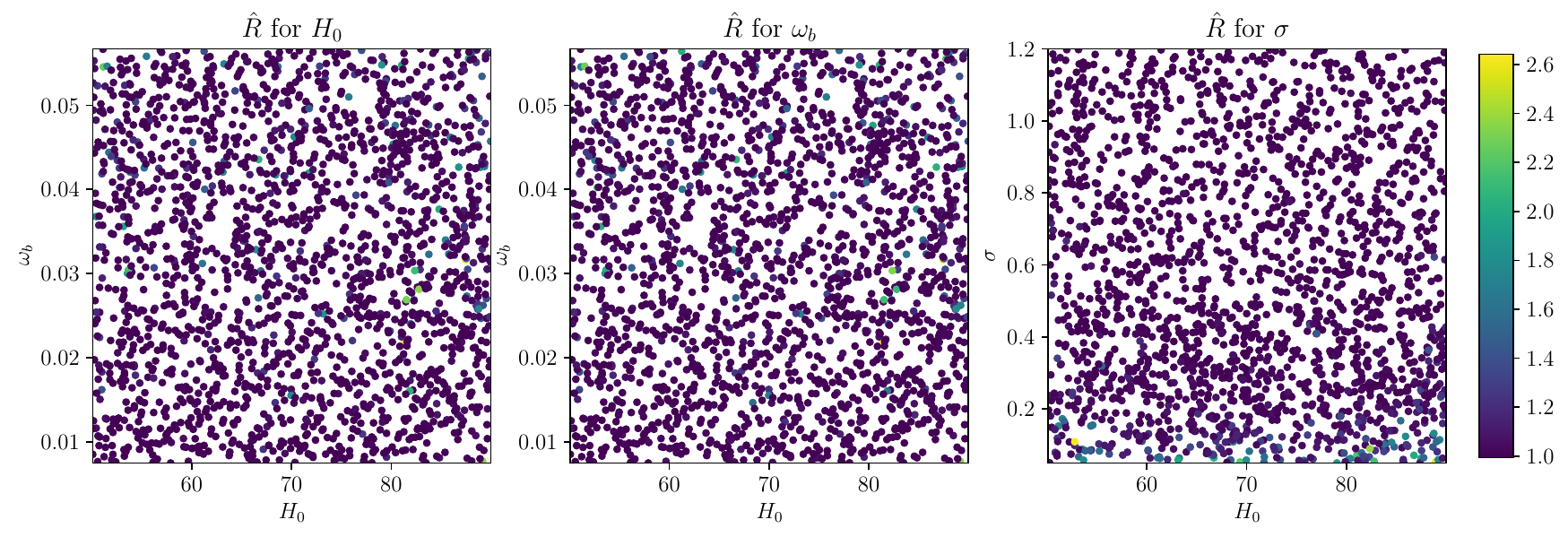}
     \caption{ESS (top) and $\hat{R}$ (bottom) statistics for inferences on dust/CMB mixtures across the parameter space.}
    \label{fig:cosmo_images_validation}
\end{figure*}

\end{document}